\pdfoutput=1
\documentclass[11pt]{article}

\usepackage[final]{acl}

\usepackage{times}
\usepackage{latexsym}
\usepackage{amsmath}
\usepackage{amsfonts} % 或者使用 \usepackage{amssymb}
\usepackage{booktabs} % for professional tables
\usepackage{multirow}
\usepackage{arydshln}
\usepackage{xcolor}
\usepackage{colortbl}
\usepackage[most]{tcolorbox}
\definecolor{darkblue}{rgb}{0.0,0.0,0.5}
\definecolor{purple}{rgb}{0.5,0.0,0.5}
\newcommand{\code}[1]{{\ttfamily#1}}
\definecolor{darkgray}{rgb}{0.4, 0.4, 0.4}
\newcommand{\db}[1]{\textcolor{darkgray}{\bf\scriptscriptstyle \selectfont \,(#1)}}
\newcommand{\dbb}[1]{\textcolor{black}{\bf\scriptscriptstyle \selectfont \,(#1)}}
\newcommand{\ddel}[1]{\textcolor{darkgray}{\rm\scriptscriptstyle \selectfont \,(#1)}}
\usepackage[T1]{fontenc}
\usepackage[utf8]{inputenc}

\usepackage{microtype}

\usepackage{inconsolata}

\usepackage{graphicx}
\usepackage{subcaption}
\usepackage{threeparttable}
\title{Improving Zero-shot LLM Re-Ranker with Risk Minimization}
\author{
Xiaowei Yuan$^{1,2,3}$, Zhao Yang$^{1,2}$, Yequan Wang$^{3,*}$, Jun Zhao$^{1,2}$, Kang Liu$^{1,2,*}$\\
$^1$The Key Laboratory of Cognition and Decision Intelligence for Complex Systems,\\
Institute of Automation, Chinese Academy of Sciences\\
$^2$School of Artificial Intelligence, University of Chinese Academy of Sciences\\
$^3$Beijing Academy of Artificial Intelligence, Beijing, China\\
\texttt{
yuanxiaowei2022@ia.ac.cn,
\{zhao.yang, jzhao, kliu\}@nlpr.ia.ac.cn}\\
\texttt{tshwangyequan@gmail.com}
}

\begin{document}
\maketitle
\renewcommand{\thefootnote}{\fnsymbol{footnote}}
\footnotetext[1]{Corresponding authors.}
\renewcommand{\thefootnote}{\arabic{footnote}}
\begin{abstract}
% Large Language Models (LLMs) have exhibited remarkable zero-shot generalization across diverse natural language processing tasks, including information retrieval. 
In the Retrieval-Augmented Generation (RAG) system, advanced Large Language Models (LLMs) have emerged as effective Query Likelihood Models (QLMs) in an unsupervised way, which re-rank documents based on the probability of generating the query given the content of a document. However, directly prompting LLMs to approximate QLMs inherently is biased, where the estimated distribution might diverge from the actual document-specific distribution.
In this study, we introduce a novel framework, $\mathrm{UR^3}$, which leverages Bayesian decision theory to both quantify and mitigate this estimation bias. Specifically, $\mathrm{UR^3}$ reformulates the problem as maximizing the probability of document generation, thereby harmonizing the optimization of query and document generation probabilities under a unified risk minimization objective. Our empirical results indicate that $\mathrm{UR^3}$ significantly enhances re-ranking, particularly in improving the Top-1 accuracy. It benefits the QA tasks by achieving higher accuracy with fewer input documents. 

\end{abstract}

\section{Introduction}
% Large Language Models (LLMs), such as ChatGPT and GPT-4 (OpenAI, 2022, 2023), are revolutionizing natural language processing with strong zero-shot and few-shot generalization. By pretraining on large-scale text corpora and alignment fine-tuning to follow human instructions, LLMs have demonstrated their superior capabilities in language understanding, generation, interaction, and reasoning (Ouyang et al., 2022).

% Large language models (LLMs) have demonstrated remarkable capabilities and significantly advanced the field of natural language processing~\cite{journals/corr/abs-2303-18223}.
% Particularly, the zero-shot generalization capabilities of Large language models (LLMs) have revolutionized the way AI systems can be applied across diverse tasks without explicit task-specific training~\cite{ChatGPT,OpenAI2023GPT4TR}.
Large Language Models (LLMs) exhibit remarkable capabilities but face several challenges including hallucination and outdated knowledge~\cite{llm-zhaoxin,journals/csur/JiLFYSXIBMF23}. Retrieval-Augmented Generation (RAG) has emerged as a promising solution by incorporating external knowledge~\cite{ram-etal-2023-context,journals/corr/abs-2312-10997}.
In the RAG system, a re-ranking model can serve as a second-pass document optimizer and refiner for the knowledge retrieval. This is particularly critical in open-domain Question Answering (QA) tasks, where it leads to large gains in performance~\cite{karpukhin-etal-2020-dense,journals/corr/abs-2308-07107}.
The re-ranker assesses the relevance of the documents retrieved by the initial retriever (e.g., BM25~\cite{BM25}) and effectively prioritizes the most relevant items at the top.
This not only enhances retrieval efficiency and responsiveness but also resolves the challenge of context window expansion by limiting the total number of documents~\cite{journals/corr/abs-2312-10997}.
% Re-ranking is the core task in IR applications, especially for the retrieval-augmented generation for AI-generated content like the open-domain question answering (QA).
% Text retrieval is a core sub-task in many NLP problems, for example, open-domain question answering where a document must be retrieved and then read to answer an input query.

\begin{figure}[t] 
  \includegraphics[width=\columnwidth]{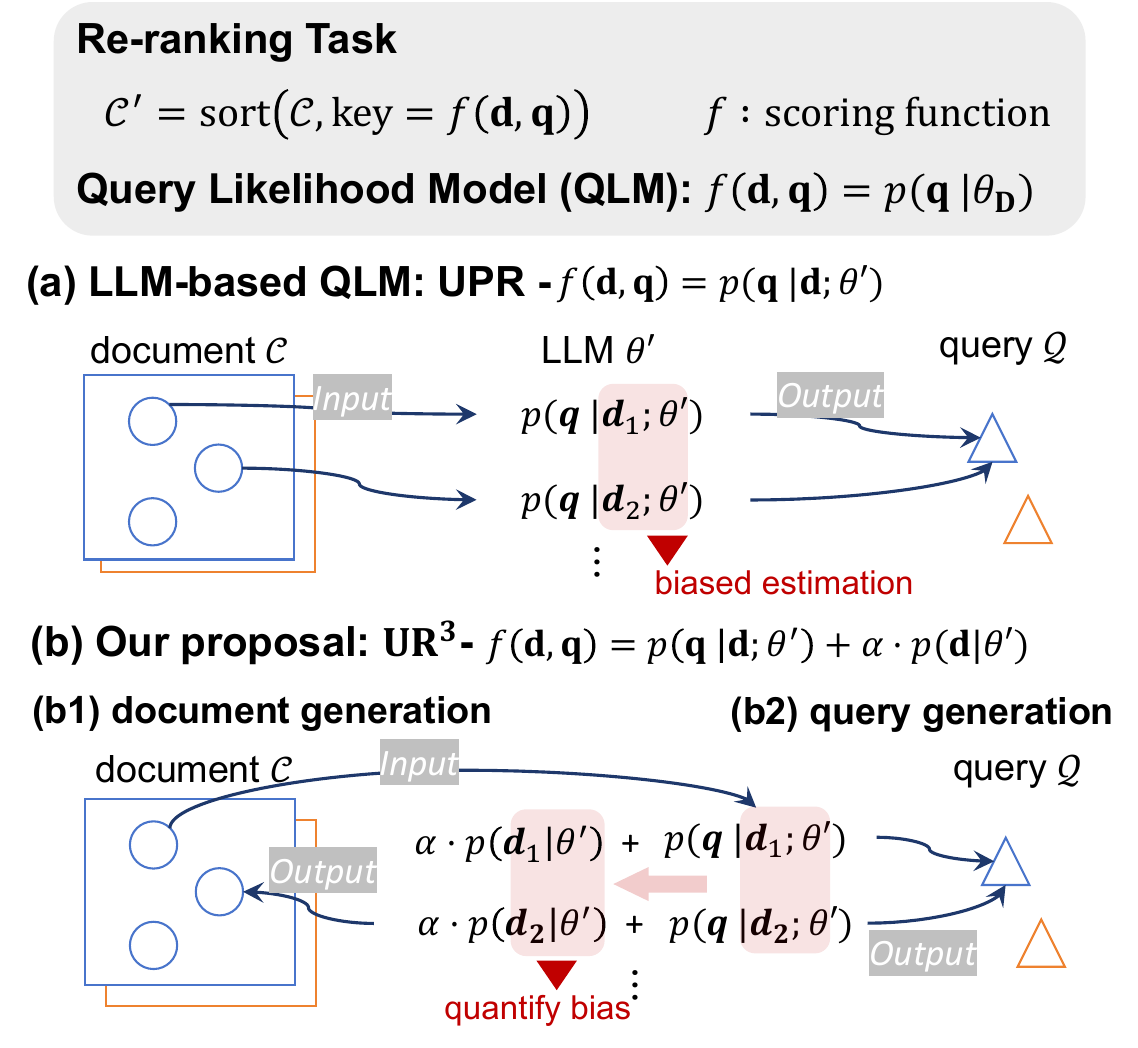}
  \caption{Method comparison in the re-ranking task. (a) The framework of LLM-based QLM method: unsupervised passage re-ranker (UPR). (b) The framework of our proposal: Unsupervised Risk-minimization Re-Ranker ($\mathbf{UR^3}$); (b1) calculating document generation probability to quantify the biased model estimation; (b2) calculating the query generation probability to measure relevance.}
  \label{fig:intro}
\end{figure}
Most previous approaches trained the re-ranker on manual supervision signals~\cite{karpukhin-etal-2020-dense,nogueira-etal-2020-document,SPLADE}, which require significant human efforts and demonstrate weak generalizability~\cite{journals/corr/abs-2112-09118, Mokrii}.
% in bi-encoder~\cite{karpukhin-etal-2020-dense}, cross-encoder~\cite{nogueira-etal-2020-document}, and sparse~\cite{SPLADE} architectures, which require significant human effort and demonstrate weak generalizability~\cite{nguyen2016ms,journals/corr/abs-2112-09118}.
As the size of models scales up (e.g., exceeding 10 billion parameters), it becomes increasingly difficult to fine-tune the dedicated re-ranking models. 
%Therefore, there is a growing interest in leveraging the zero-shot language understanding and reasoning capabilities of LLMs in the IR area.
To address this challenge, recent efforts have attempted to leverage the zero-shot language understanding and generation capabilities of LLMs to directly enhance document re-ranking in an unsupervised way.

Recent studies have explored LLMs for permutation generation~\cite{journals/corr/abs-2305-02156,conf/emnlp/0001YMWRCYR23} as re-rankers, which yield significant performance by generating a ranked list of a group of documents. However, these models face high time complexity with long lists and the performance is highly sensitive to the document order in the prompt.~\cite{journals/corr/abs-2308-07107}.
% and only the GPT-4~\cite{OpenAI2023GPT4TR} based method can achieve competitive performance.
In this paper, we consider a unsupervised query generation method based on Query Likelihood Model (QLM)~\cite{conf/sigir/PonteC98,phd/basesearch/Hiemstra01,conf/sigir/ZhaiL01}, which judges the relevance of each query-document pair independently, thus offering lower time complexity. The core idea behind QLM is to infer a language model $\theta_D$ for each document $\mathbf{d}$, and to rank the documents based on the likelihood of the query according to this model $p(\mathbf{q}\mid \theta_D)$. 
% The pointwise method judges the relevance of each query-document pair independently, thus offering lower time complexity and enabling batch inference.
% The intuition is to capture which documents best "fit" the particular query $\mathbf{q}$, thereby representing a notion of $relevance$. 

The typical LLM-based QLM is called Unsupervised Passage Re-ranker (UPR)~\cite{conf/emnlp/SachanLJAYPZ22}. It leverages a LLM $\theta'$ to score the probability of generating the question $\mathbf{q}$ conditioned on the input document $\mathbf{d}$ as $p(\mathbf{q} \mid \mathbf{d}; \theta')$, highlighting the zero-shot ranking capabilities of the LLM-based QLM.
Upon closer examination, \textbf{an inherent estimation bias occurs} when employing $p(\mathbf{d};\theta')$ to approximate $p(\theta_D)$.
As illustrated in Figure~\ref{fig:intro}, the estimated distribution $p(\mathbf{d};\theta')$ might not accurately reflect the actual document-specific distribution, $p(\theta_D)$. This divergence primarily stems from the estimation bias in employing a generalized model, such as $\theta'$, which is not specifically tuned to capturing the document characteristics necessary for the query generation task~\cite{conf/acl/BenderK20,conf/acl/WangXFLSX0022,journals/corr/abs-2302-10198}.

To bridge the gap between the estimated distribution by LLM $p(\mathbf{q} \mid \mathbf{d}; \theta')$ and the actual document distribution $p(\theta_D)$, we introduce a novel method called \underline{U}nsupervised \underline{R}isk-minimization \underline{R}e-\underline{R}anker ($\mathbf{UR^3}$). 
It characterizes the document selection as a optimization process based on Bayesian decision theory~\cite{books/wi/Wald50,zhai2006risk}. 
In specific, to quantify the estimation bias,
$\mathrm{UR^3}$ employs the Kullback-Leibler (KL) divergence~\cite{kl} to reformulate the minimization of bias as the maximization of document generation probability. Therefore, this approach allows for the simultaneous maximization of both query and document generation probabilities, treating them as a common objective in term of risk minimization.

To prove the eﬀectiveness of $\mathrm{UR^3}$, we verify it in the re-ranking stage in current RAG models for the open-domain QA tasks.
% In comparison with UPR and other baselines. 
In the re-ranking tasks, the results indicate that our method significantly enhances the Top-1 accuracy on the open-domain NQ~\cite{journals/tacl/KwiatkowskiPRCP19}, WebQ~\cite{conf/emnlp/BerantCFL13}, and TriviaQA~\cite{conf/acl/JoshiCWZ17} datasets, with improvements of 6.64\%, 6.35\%, and 3.18\% points compared with UPR. In the QA tasks, the Exact Match (EM) and F1 scores exhibit increases of up to 1.48 and 2.06, respectively, when utilizing the fewest document input (only 1).
% Additionally, the nDCG@20 and MAP@100 metrics show increases of up to 2.88\% and 2.74\%, respectively.
% can effectively improve the top-tier document ranking both on the Open Question Answering (QA) and retrieval tasks.

The contributions of this paper are as follows:
\begin{itemize}
    \item From the perspective of risk minimization, this paper presents a theoretical formalization to rank the relevance of query-document pairs. This formalization not only considers query generation but also evaluates the estimation bias through document generation probabilities (See \S\ref{exp:rerank}).
    % \item the distribution bias when using LLM as a QLM model is quantified, effectively enhancing performance in diverse retrieval tasks. % with 可解释分析
    \item The enhancement in performance is notable for higher-ranked results, with the most pronounced improvements at the Top-1. This significantly benefits the QA tasks by achieving higher accuracy with fewer input documents (See \S\ref{exp:qa}).
    % Particularly, we demonstrate that, when employing a generation model from the same series with the re-ranker, the performance in QA task can be more enhanced 
    % (See \S\ref{exp:analysis}).
\end{itemize}

\section{Related Work}
Re-rankers serve as the second-pass document filter in IR, 
% designed to re-rank a list of documents initially retrieved by a primary retriever (e.g., BM25), 
based on the relevance between the query and the documents. Recently, LLMs have attracted significant attention in the field of IR, with numerous innovative approaches being proposed for re-ranking tasks~\cite{journals/corr/abs-2308-07107,journals/corr/abs-2312-10997}.
Existing instructions for zero-shot document re-ranking with LLMs can be classified into three types: query generation~\cite{sachan-etal-2021-end,conf/emnlp/Zhuang0KZ23}, relevance generation~\cite{journals/corr/abs-2211-09110} and permutation generation~\cite{journals/corr/abs-2305-02156,conf/emnlp/0001YMWRCYR23}. However, permutation generation 
models face high time complexity with long lists, and relevance generation method does not have an advantage in terms of performance compared to others~\cite{journals/corr/abs-2308-07107}.
In this paper, we focus on the application of query generation LLMs in an unsupervised way.
 
% \subsection{Query Generation with LLMs}
Language modeling approaches to information retrieval are attractive and promising because they connect the problem of retrieval with that of language model estimation.
UPR~\cite{conf/emnlp/SachanLJAYPZ22} introduces instructional query generation methods by LLMs, as the query-document relevance score is determined by the average log-likelihood of generating the actual query tokens based on the document. It has been proven that some LLMs yield significant performance in zero-shot document re-ranking. Recently, research~\cite{conf/emnlp/Zhuang0KZ23} has also shown that the LLMs that are pre-trained without any supervised instruction fine-tuning (such as LLaMA~\cite{llama1}) also yield robust zero-shot ranking ability.

Another line is to optimize prompt for better performance. For example, a discrete prompt optimization method Co-Prompt~\cite{co-prompt} is proposed for better prompt generation in re-ranking tasks. Besides, PaRaDe~\cite{parade} introduces a difficulty-based method for selecting few-shot demonstrations to include in prompts, demonstrating significant improvements over zero-shot prompts.
But the prompt engineering is not within the scope of this paper. Our prompt adheres to the original setup as UPR (e.g., "\textit{Please write a query based on this document}") in a zero-shot manner. 
% The detailed instructions are included in Appendix~\ref{sec:prompt}.
% \subsection{Relevance and Permutation generation}

% Zhuang et al. [147] propose to incorporate fine-grained relevance labels (e.g., “highly relevant”, “somewhat rele- vant” and “not relevant”) into the prompt, which helps LLMs more effectively differentiate among documents with varying levels of relevance to a query.

\section{$\mathbf{UR^3}$: Unsupervised Risk-minimization Re-Ranker}
% In a retrieval system, the fundamental task involves presenting a sequence of documents to the user. The choice of which documents to retrieve for ranking is based on their $relevance$ to the user's query. Here we focus on second-stage retrieval, which re-ranks a set of candidate documents based on a off-the-shelf retriever.

Existing methods~\cite{conf/emnlp/SachanLJAYPZ22, conf/emnlp/Zhuang0KZ23} have limited performance in re-ranking due to the oversight of biased estimation when considering a LLM conditioned on the input document $p(\mathbf{d};\theta')$ as the actual document language distribution $p(\theta_D)$ . 

To tackle the problem, we introduce a novel re-ranking model $\mathbf{UR^3}$, which considers not only the query generation probability (\S \ref{sec:qg}) but also the quantification of bias (\S \ref{sec:dg}). For the latter, our method characterizes the distribution discrepancy between an actual document language model $p(\theta_D)$ and the LLM $p(\mathbf{d};\theta')$. Utilizing the distance-based risk-minimization Bayes decision, the estimation bias can be reformulated as the probability of document generation, thereby forming a common optimization objective with the query generation process.

% Figure~\ref{fig:intro} illustrates .. %introduction fig

\subsection{Problem Formalization}
In a retrieval system, a query $\mathbf{q}$ from a user $\mathcal{U}$ is assumed to sampled from a query-based empirical distribution $p(\mathbf{q} \mid \theta_{Q_e})$\footnote{Here we do not define a user-specific query model that encodes detailed knowledge about the user, but rather an empirical distribution $\theta_{Q_e}$ for mathematical convenience. The query language model is concentrated on the actual query terms.}.
A document model $\theta_D$ is selected from the document source $\mathcal{S}$ according to the distribution $p(\theta_D \mid \mathcal{S})$, and then this model generates a document according to $p (\mathbf{d} \mid \theta_D)$. 

% The fundamental task involves presenting a sequence of documents to the user. 
% The choice of which documents to retrieve for ranking is based on their $relevance$ to the user's query.  
% Here we focus on second-stage retrieval, which re-ranks a set of candidate documents based on a off-the-shelf retriever.
Let $\mathcal{C} = \{\mathbf{d}_1, \mathbf{d}_2,...,\mathbf{d}_k\}$ be a set of candidates from source $\mathcal{S}$, where we assume that the retriever provides the $K$ most relevant documents. 
For a candidate document $\mathbf{d}$, QLM~\cite{conf/sigir/PonteC98} estimates the conditional probability $p(\mathbf{q}|\theta_{D})$\footnote{For convenience, the subscript $i$ is omitted in subsequent notations.}, which captures how well the document “fits” the particular query.
% The probability that document model $\theta_{D_i}$ generates the query $\mathbf{q}$.
Previous QLM-based work~\cite{conf/emnlp/SachanLJAYPZ22,co-prompt,parade} score each document by computing the likelihood of the query conditioned on the input document as $p(\mathbf{q}\mid\mathbf{d}; \theta')$. 
They approximate the document language model $\theta_{D}$ by applying $\mathbf{d}$ as input into a pre-trained LLM $\theta'$, formulated as:
\begin{equation} \label{eq:defllm}
    p(\theta'_{D}) \stackrel{\text{def}}{=} p(\mathbf{d}; \theta')
\end{equation}

\begin{figure}[t] 
  \includegraphics[width=\columnwidth]{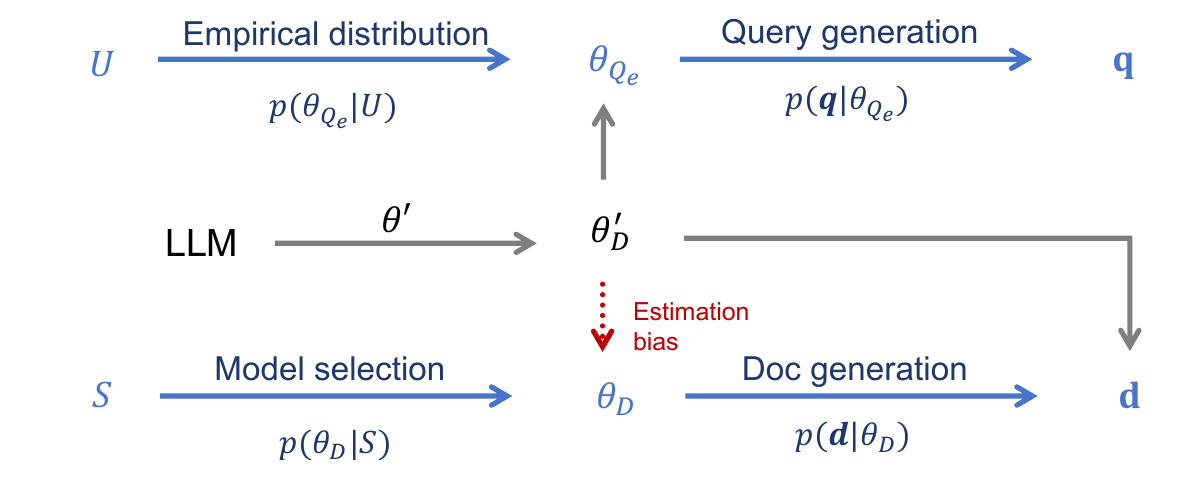}
  \caption{The process for a LLM-based re-ranking method in the view of Bayes decision theory.}
  \label{fig:framework}
\end{figure}
% The goal is to minimize the KL divergence, thereby aligning the distributions in the view of probabilistic modeling.
\subsection{Bayes Decision Theory}\label{sec:bayes}
The standard retrieval problem can be regarded as a decision problem where the decision involves choosing the best ranking. 
% We formalize this decision-theoretic view of retrieval within the framework of Bayesian decision theory.
\citet{journals/ipm/ZhaiL06} formalizes the decision problem within a probabilistic framework of the Bayesian decision theory. A possible action $a$ is to return a single document based on the expected risk $R$, which is associated with a loss $L(a,\theta)$:
\begin{equation}
    R(a\mid \mathcal{U},\mathbf{q},\mathcal{S},\mathcal{C})=\int_{\Theta} L(a,\theta)p(\theta \mid \mathcal{U},\mathbf{q},\mathcal{S},\mathcal{C}) d\theta
\end{equation}

The Bayesian decision rule is then to present the document list $a^*$ having the least expected risk:
\begin{equation}
    a^*=\underset{a}{\operatorname{arg\,min}} R(a|U,\mathbf{q},S,C)
\end{equation}

% measures the relevance score of each query-document pair by the distance between the query and document models and formalize the retrieval framework by the Bayesian decision theory. 
We extend the framework to allow for a consideration of the the approximate document model $\theta'_{D}$ with LLM $\theta'$, as illustrated in Figure~\ref{fig:framework}. The \textbf{expected risk of action $a$} can be formulated as:
\begin{align} \label{eq:action}
    R(\mathbf{d}; \mathbf{q}) &\stackrel{\text{def}}{=} R(a = \mathbf{d} \mid U, \mathbf{q}, \mathcal{S}, C, \theta') \\
    & \propto L(\hat\theta_\mathbf{q}, \theta'_D,\hat\theta_\mathbf{d}) \nonumber
\end{align}
where the distribution of $\theta'_D$ is determined by the distribution $p(\mathbf{d};\theta')$, and
% probability in Equation \ref{eq:defllm}\footnote{Here we reformulate $\theta'_D$ as $\hat\theta'_\mathbf{d}$ to keep consistency with the subsequent notation.}, and
\begin{align*}
&\hat{\theta}_\mathbf{q} = \underset{\theta_{Q_e}}{\operatorname{arg\,max}} p(\theta_{Q_e} \mid \mathbf{q}, \mathcal{U}) \\
&\hat{\theta}_\mathbf{d} = \underset{\theta_D}{\operatorname{arg\,max}} p(\theta_D \mid \mathbf{d}, \mathcal{S}) 
\end{align*}
The detailed derivations are presented in Appendix~\ref{appendix:bayes}.

To summarize, the document set $\mathcal{C}$ is represented through a series of $k$ sequential decisions. This process yields a list of documents ranked in ascending order according to the $R(\mathbf{d}; \mathbf{q})$. A smaller loss $L$ means a better ranking for the document.
% Formula \ref{eq:risk} is presented in a general manner for risk minimization.

\subsection{Distance-based Loss Functions}
In this section, we conceptualize the loss function, $L$, as a distance-based function, $\Delta$, quantified using KL divergence, initially introduced by~\citet{conf/sigir/LaffertyZ01}. 
% Although KL divergence does not represent a true metric distance between distributions due to its asymmetry and non-compliance with the triangle inequality, it is frequently beneficial to consider it as a "distance" measure between distributions~\cite{books/daglib/0016881}.  
% This allows for a united optimization problem for measuring different distributions of language models.
% equivalently reformulate the loss function $L$ by a KL distanced-based measure  to derive the final probabilistic distance re-ranking model.

% Based on the distanced-based loss function $L$ on KL model~\cite{conf/sigir/LaffertyZ01}, we derive a general probabilistic distance re-ranking model.
% Formally, let $L$ be proportional to a distance measure $\Delta$ among models. 
Based on the dependency relationships illustrated in Figure~\ref{fig:framework}, the distance among models can be split into the sum of the following two terms, where the details refer to Appendix~\ref{appendix:loss}.
% \footnote{The theoretical derivations are detailed in Appendix~\ref{appendix:derivation}.}\label{commonfootnote}:
\begin{equation}
 L(\hat\theta_\mathbf{q}, \theta'_D,\hat\theta_\mathbf{d})
 % & \stackrel{\text{def}}{=} c \Delta (\hat\theta_\mathbf{q}, \theta'_D,\hat\theta_\mathbf{d}) \\
\approx c_1\Delta(\hat\theta_\mathbf{q}, \theta'_D) +c_2\Delta(\theta'_D, \hat\theta_\mathbf{d}) 
\end{equation}
where $c_1>0$ and $c_2>0$ are constants. 
Therefore, the following formula can be derived:
\begin{equation}
% R(\mathbf{d}; \mathbf{q}) &\propto \Delta(\hat{\theta}_\mathbf{q}, \hat{\theta'}_\mathbf{d},\hat{\theta}_\mathbf{d}) \\
R(\mathbf{d}; \mathbf{q})\propto \Delta(\hat{\theta}_\mathbf{q}, \theta'_D) + \alpha \Delta(\hat{\theta}_\mathbf{d}, \theta'_D) 
\end{equation}
where the $\alpha$ is proportional to $c_2 /\ c_1$. Then we will characterize that the minimum risk ranking criterion as the sum of probability of query generation (\S \ref{sec:qg}) and document generation (\S \ref{sec:dg}), respectively.
\subsubsection{Probability of Query Generation} \label{sec:qg}
Given $\hat{\theta}_\mathbf{q}$ is a distribution that represents an empirical distribution of query $\mathbf{q}$, where $\mathbf{q} = q_1q_2\ldots q_m$, we have\footnote{The theoretical derivations are detailed in Appendix~\ref{appendix:derivation}.}:
% \footnotemark[\value{footnote}]:
% Detailed derivation refers to~\citet{conf/sigir/LaffertyZ01}
\begin{align} \label{eq:kl1}
\Delta(\hat\theta_\mathbf{q}, \hat\theta'_\mathbf{d}) &\stackrel{\text{def}}{=} \text{KL}[\ p(\hat\theta_\mathbf{q}) \parallel p(\theta'_D)\ ] \\
&\propto - \log p(\mathbf{q} \mid \theta'_D) + c_\mathbf{q} \nonumber\\
&\propto -\frac{1}{m}\sum_{i=1}^{m} \log p(q_i \mid \mathbf{q}_{<i}, \mathbf{d}; \theta') \nonumber
\end{align}
where the constant $c_\mathbf{q}$ presents the entropy of the query model.
This is precisely the log-likelihood criterion that has been in used in the language modeling approaches of query generation~\cite{conf/emnlp/SachanLJAYPZ22,conf/emnlp/Zhuang0KZ23}.

\subsubsection{Probability of Document Generation} \label{sec:dg}
% where \(p(\theta'_d)\) denotes the distribution under the language model \(\theta'\) with \(d\) as the input, 
Following previous studies~\cite{conf/sigir/PonteC98,phd/basesearch/Hiemstra01,conf/sigir/ZhaiL01}, $p(\mathbf{d})$ is assumed to be uniformly distributed, if we view $\theta'$ as a stochastic variable, then 
\begin{equation}
    P(\mathbf{d}, \theta') = P(\mathbf{d})P(\theta' \mid \mathbf{d}) \propto P(\theta' \mid \mathbf{d})
\end{equation}
Therefore, the distance from the approximate distribution $\theta'_D$ to the actual posterior distribution $\hat\theta_\mathbf{d}$ is formulated as:
\begin{align} \label{eq:Delta}
\Delta(\hat\theta_\mathbf{d}, \theta'_D) &\stackrel{\text{def}}{=} \text{KL}[\ p(\hat\theta_\mathbf{d}) \parallel p(\mathbf{d}, \theta')\ ] \\
&\propto \text{KL}[\ p(\hat\theta_\mathbf{d}) \parallel p(\theta' \mid \mathbf{d})\ ] \nonumber
\end{align}

The calculation of Formula~\ref{eq:Delta} can be equivalently reformulated as the computation of the Evidence Lower Bound (ELBO) via variational inference~\cite{journals/jmlr/HoffmanBWP13}\footnotemark[\value{footnote}]:
\begin{equation}
    \text{KL}[\ p(\hat\theta_\mathbf{d}) \parallel p(\theta' \mid \mathbf{d})] = -\text{ELBO}(\theta) + \log p(\mathbf{d}) 
\end{equation}
where
\begin{equation*}
    \text{ELBO}(\theta) = \mathbb{E}[\log p(\mathbf{d} \mid \theta')] - \text{KL}[\ p(\hat\theta_\mathbf{d}) \parallel p(\theta')] 
\end{equation*}
Since the latter KL divergence term in $\text{ELBO}(\theta)$ is same for all $\mathbf{d}$ for a specific LLM, the following formula can be derived:
\begin{align} \label{eq:kl2}
    \Delta(\hat\theta_\mathbf{d}, \hat\theta'_\mathbf{d}) & \propto -\mathbb{E}[\log p(\mathbf{d} \mid \theta')] 
    % &\propto - \log p(\mathbf{d} \mid \theta') \nonumber
\end{align}

Let $\mathbf{d}=d_1, d_2, ..., d_n$, the final risk minimization object can be formulated as the proportional sum of query and document generation probabilities based on Formula~\ref{eq:kl1} and~\ref{eq:kl2}: 
\begin{align}\label{eq:total}
    R(\mathbf{d}; \mathbf{q}) &\propto -\frac{1}{m}\sum_{i=1}^{m} \log p(q_i \mid \mathbf{q}_{<i}, \mathbf{d}; \theta') \\
    & - \alpha \cdot\left(\frac{1}{n} \sum_{i=1}^{n} \log p(d_i \mid\mathbf{d}_{<i};\theta')\right) \nonumber
\end{align}
where $\alpha$ is a hyperparameter. The expectation of the term in Formula~\ref{eq:kl2} is calculated as the document generation probability on LLM $\theta'$, which synchronizes the computation of the query and the document within one-time inference. The detailed instructions are included in Appendix~\ref{sec:prompt}.
% to minimize computational costs.

\section{Experiments}
\subsection{Experimental Setup}
% In this section, we describe the datasets, unsupervised and supervised retrievers, and baseline methods used for our document re-ranking experiments.

\paragraph{Datasets.}
For the document retrieval in the QA task, we use the three popular datasets of open-domain QA: NaturalQuestions (NQ;~\cite{journals/tacl/KwiatkowskiPRCP19}), WebQuestions (WebQ;~\cite{conf/emnlp/BerantCFL13}) and  TriviaQA~\cite{conf/acl/JoshiCWZ17}. For re-ranking, we utilize the pre-processed English Wikipedia dump from December 2018, as released by~\citet{karpukhin-etal-2020-dense}, as the source of evidence documents. Then we apply the ranking results to generate answers for questions to evaluate the QA performance.

We additionally employ the BEIR Benchmark~\cite{conf/nips/Thakur0RSG21} for a comprehensive retrieval evaluation in Appendix~\ref{appendix:beir}.
% This benchmark encompasses a variety of datasets, each containing test set queries, evidence documents, and annotations denoting document relevance.

\paragraph{Retrievers.}
In our re-ranking experiments, we retrieve documents from both unsupervised and supervised retrievers, including three unsupervised retrievers—Contriever~\cite{journals/corr/abs-2112-09118}, BM25~\cite{BM25}, and MSS~\cite{sachan-etal-2021-end}—and one supervised retriever, DPR~\cite{karpukhin-etal-2020-dense}.
% BM25 ranks based on the term-frequency and inverse document frequency of the keywords present in the question and passage. 
% Contriever uses momentum contrastive training to learn dense retrievers from text paragraphs. Such training has shown to obtain strong zero-shot retrieval performance on many benchmarks.
% MSS is a dense retriever trained by predicting masked salient spans like named entities with the help of a reader network. MSS pre-training has also shown to improve supervised retrieval performance.
% DPR uses annotated question-context paragraphs and hard negative examples to train a supervised dense retriever.

\paragraph{Baselines}
We adopt three unsupervised re-ranking methods as the baselines: RankGPT (RG)~\cite{conf/emnlp/0001YMWRCYR23}, UPR~\cite{conf/emnlp/SachanLJAYPZ22} and Interpolation (Int.)~\cite{conf/emnlp/Zhuang0KZ23}. 
\begin{itemize}
    \item RankGPT aims to directly rank a list of documents employing a sliding window strategy to re-rank subsets of candidate documents based on a LLM\footnote{For a fair comparison, the implementation of the RankGPT method is based on the LLaMA2-7B-Chat model.}.
    \item UPR leverages a LLM to obtained the query-document relevance score, which is determined by the log-likelihood of generating the actual query tokens based on the document.
    \item Interpolation method linearly combines the UPR score with the scores from the first-stage retriever using a weighted sum of scores. We apply the method to both UPR and $\mathrm{UR^3}$ methods with the same weight configuration. 
\end{itemize}

For $\mathrm{UR^3}$, the values of $\alpha$ is set to 0.25. Detailed analyses about the hyperparameter are provided in Appendix~\ref{appendix:hyper}.

\paragraph{Metrics}
Following previous work~\cite{conf/nips/Thakur0RSG21,conf/emnlp/SachanLJAYPZ22}, we compute the conventional Top-K retrieval accuracy, nDCG@K and MAP@K metrics to evaluate the re-ranking performance. 
% Top-K accuracy is defined as the fraction of questions for which at least one passage within the top-K documents is relevant. The nDCG@K metric measures the gain from retrieving relevant documents, weighted by their position in the top K list. MAP assesses the precision of the retrieval system at each point a relevant document is retrieved. 
And we use the EM and F1 scores for evaluating the QA performance of LLMs.

\paragraph{LLMs}
For the re-ranking task, our experiments are conducted on LLaMA2 (7B)~\cite{llama2}, Mistral (7B)~\cite{mistral7b} and GPT-Neo (2.7B)~\cite{neo} models. 

For the QA task, a reader processes the documents retrieved by the retriever to generate the answer to the query. We respectively employ the LLaMA2 (7B and 13B), Mistral (7B) and Gemma (7B)~\cite{gemma} models as the reader. 
% An open-domain QA system is typically composed of two primary components: a retriever and a reader. 

% During re-ranking and QA, outputs are generated using greedy decoding.
% For the QA task, we conduct experiments on LLaMA2 (13B), Mistral (7B) and Gemma (7B) models.

\addtolength{\tabcolsep}{-0.3em}
\begin{table*}[t]
    \tiny
    \centering
    \resizebox{\textwidth}{!}{%
    \begin{threeparttable}
    \begin{tabular}{llcccc|cccc|cccc|cccc}
         \cmidrule[1pt]{1-18}
         % & & \multicolumn{9}{c}{\textit{Unsupervised Retrievers}} & \multicolumn{3}{c}{\textit{Supervised Retrievers}} \\
         & & \multicolumn{4}{c}{\textbf{Contriever}} & \multicolumn{4}{c}{\textbf{BM25}} & \multicolumn{4}{c}{\textbf{MSS}} & \multicolumn{4}{c}{\textbf{DPR}}  \\
         \cmidrule(lr){3-6} \cmidrule(lr){7-10} \cmidrule(lr){11-14} \cmidrule(lr){15-18} %\cmidrule(lr){15-17}
         \textbf{Datasets} &\textbf{Metric} & Orig.* & RG & UPR$\db{+Int.}$ & $\mathrm{UR^3}\db{+Int.}$ &  Orig. & RG & UPR$\db{+Int.}$ &  $\mathrm{UR^3}\db{+Int.}$ &  Orig. & RG & UPR$\db{+Int.}$ &   $\mathrm{UR^3}\db{+Int.}$ &  Orig. & RG & UPR$\db{+Int.}$ & $\mathrm{UR^3}\db{+Int.}$  \\
         \midrule
         \multirow{7}{*}{NQ}& Top-1 & \cellcolor{gray!10}22.16 & 13.07 & 32.38$\ddel{32.49}$ & \cellcolor{blue!4}\textbf{37.67}$\ddel{36.37}$ & \cellcolor{gray!10}22.11 & 17.51 & 32.69$\ddel{33.10}$ & \cellcolor{blue!4}\textbf{38.01}$\ddel{37.42}$ & \cellcolor{gray!10}19.28 & 15.35 & 32.83$\ddel{33.49}$ & \cellcolor{blue!4}\textbf{37.48}$\ddel{36.43}$ & \cellcolor{gray!10}\textcolor{red}{46.34} & 37.06 & 37.65$\ddel{48.45}$ & \cellcolor{blue!4}44.29$\dbb{52.24}$  \\
        % & Top-3 & \cellcolor{gray!10}36.59 & 52.11 & 54.74 & \cellcolor{gray!10}36.59 & 52.11 & 54.74 & \cellcolor{gray!10}0.91 & 0.88 & 0.84 & \cellcolor{gray!10}0.78 & 0.71 & 0.82 \\
        & Top-5 & \cellcolor{gray!10}47.29 & 46.87 & 61.41$\ddel{61.00}$ & \cellcolor{blue!4}63.96$\dbb{64.10}$ & \cellcolor{gray!10}43.77 & 38.25 & 59.83$\ddel{59.50}$ & \cellcolor{blue!4}\textbf{61.97}$\ddel{61.19}$ & \cellcolor{gray!10}41.25 & 35.76 & 59.28$\ddel{59.22}$ & \cellcolor{blue!4}\textbf{61.08}$\ddel{60.61}$ & \cellcolor{gray!10}68.28 & 67.67 & 69.20$\ddel{73.85}$ & \cellcolor{blue!4}71.99$\dbb{74.99}$ \\
        % & Top-10 & \cellcolor{gray!10}54.46 & 67.26 & 68.50 & \cellcolor{gray!10}54.46 & 67.26 & 68.50 & \cellcolor{gray!10}0.91 & 0.88 & 0.84 & \cellcolor{gray!10}0.78 & 0.71 & 0.82 \\
        & Top-20 & \cellcolor{gray!10}67.87 & 67.51 & 76.12$\ddel{76.26}$ & \cellcolor{blue!4}\textbf{76.57}$\ddel{76.59}$ & \cellcolor{gray!10}62.94 & 62.94 & \textbf{73.16}$\ddel{72.63}$ & \cellcolor{blue!4}72.96$\ddel{72.88}$ & \cellcolor{gray!10}59.97 & 60.22 & 71.30$\ddel{70.97}$ & \cellcolor{blue!4}\textbf{71.47}$\ddel{71.25}$ & \cellcolor{gray!10}80.06 & 79.70 & 82.66$\ddel{83.10}$ & \cellcolor{blue!4}82.99$\dbb{83.32}$ \\
        % & Top-50 & \cellcolor{gray!10}0.86 & 0.86 & 0.86 & \cellcolor{gray!10}72.74 & 76.87 & 76.87 & \cellcolor{gray!10}0.91 & 0.88 & 0.84 & \cellcolor{gray!10}0.78 & 0.71 & 0.82 &  \\
        \cmidrule(lr){2-18}
        & nDCG@1 & \cellcolor{gray!10}22.16 & 13.07 & 32.38$\ddel{32.49}$ & \cellcolor{blue!4}\textbf{37.67}$\ddel{36.37}$ & \cellcolor{gray!10}22.11 & 17.51 & 32.69$\ddel{33.10}$ & \cellcolor{blue!4}\textbf{38.01}$\ddel{37.42}$ & \cellcolor{gray!10}19.28 & 15.35 & 32.83$\ddel{33.39}$ & \cellcolor{blue!4}\textbf{37.48}$\ddel{36.43}$ & \cellcolor{gray!10}\textcolor{red}{46.34} & 17.06 & 37.65$\ddel{48.45}$ & \cellcolor{blue!4}44.29$\dbb{52.24}$  \\
        % & nDCG@3 & \cellcolor{gray!10}21.56 & 32.90 & 36.83 & \cellcolor{gray!10}21.56 & 32.90 & 36.83 & \cellcolor{gray!10}0.91 & 0.88 & 0.84 & \cellcolor{gray!10}0.78 & 0.71 & 0.82  \\
        & nDCG@5 & \cellcolor{gray!10}21.70 & 19.10 & 33.35$\ddel{33.08}$ & \cellcolor{blue!4}\textbf{36.89}$\ddel{36.36}$ & \cellcolor{gray!10}21.63 & 17.43 & 33.89$\ddel{33.89}$ & \cellcolor{blue!4}\textbf{37.12}$\ddel{36.51}$ & \cellcolor{gray!10}18.97 & 15.38 & 34.39$\ddel{34.45}$ & \cellcolor{blue!4}\textbf{37.12}$\ddel{36.53}$ & \cellcolor{gray!10}40.62 & 32.79 & 38.94$\ddel{45.51}$ & \cellcolor{blue!4}43.05$\dbb{47.97}$ \\
        % & nDCG@10 & \cellcolor{gray!10}23.10 & 35.76 & 38.27 & \cellcolor{gray!10}23.10 & 35.76 & 38.27 & \cellcolor{gray!10}0.91 & 0.88 & 0.84 & \cellcolor{gray!10}0.78 & 0.71 & 0.82 \\
        & nDCG@20 & \cellcolor{gray!10}26.15 & 24.20 & 39.08$\ddel{38.79}$ & \cellcolor{blue!4}\textbf{41.60}$\ddel{41.26}$ & \cellcolor{gray!10}25.75 & 23.45 & 39.27$\ddel{39.17}$ & \cellcolor{blue!4}\textbf{41.27}$\ddel{40.87}$ & \cellcolor{gray!10}22.88 & 21.10 & 39.36$\ddel{39.18}$ & \cellcolor{blue!4}\textbf{41.15}$\ddel{40.36}$ & \cellcolor{gray!10}42.42 & 36.43 & 44.78$\ddel{49.34}$ & \cellcolor{blue!4}47.66$\dbb{50.95}$  \\
        % & nDCG@50 & \cellcolor{gray!10}25.75 & 39.70 & 41.15 & \cellcolor{gray!10}32.48 & 44.81 & 46.69 & \cellcolor{gray!10}0.91 & 0.88 & 0.84 & \cellcolor{gray!10}0.78 & 0.71 & 0.82 \\
        % & nDCG@100 & \cellcolor{gray!10}25.75 & 39.70 & 41.15 & \cellcolor{gray!10}40.12 & 49.27 & 50.94 & \cellcolor{gray!10}0.91 & 0.88 & 0.84 & \cellcolor{gray!10}0.78 & 0.71 & 0.82  \\
        \cmidrule(lr){2-18}
        & MAP@100 & \cellcolor{gray!10}20.71 & 18.68 & 31.56$\ddel{31.18}$ & \cellcolor{blue!4}\textbf{33.94}$\ddel{33.46}$ & \cellcolor{gray!10}20.78 & 18.37 & 32.13$\ddel{32.05}$ & \cellcolor{blue!4}\textbf{34.05}$\ddel{33.66}$ & \cellcolor{gray!10}18.11 & 16.27 & 32.32$\ddel{31.98}$ & \cellcolor{blue!4}\textbf{34.10}$\ddel{33.23}$ & \cellcolor{gray!10}34.89 & 28.35 & 36.64$\ddel{41.39}$ & \cellcolor{blue!4}39.38$\dbb{42.91}$  \\
         % \bottomrule
        \toprule[0.5pt]
         \multirow{7}{*}{WebQ}& Top-1 & \cellcolor{gray!10}19.98 & 18.65 & 26.62$\ddel{28.05}$ & \cellcolor{blue!4}\textbf{32.53}$\ddel{30.81}$ & \cellcolor{gray!10}18.90 & 17.32 & 27.56$\ddel{28.54}$ & \cellcolor{blue!4}\textbf{33.91}$\ddel{33.56}$ & \cellcolor{gray!10}11.66 & 11.96 & 26.38$\ddel{25.44}$ & \cellcolor{blue!4}\textbf{29.38}$\ddel{27.66}$ & \cellcolor{gray!10}\textcolor{red}{44.83} & 37.16 & 39.32$\ddel{46.26}$ & \cellcolor{blue!4}42.18$\dbb{48.03}$ \\
        % & Top-3 & \cellcolor{gray!10}0.86 & 0.86 & 0.86 & \cellcolor{gray!10}0.82 & 0.77 & 0.83 & \cellcolor{gray!10}0.91 & 0.88 & 0.84 & \cellcolor{gray!10}0.78 & 0.71 & 0.82\\
        & Top-5 & \cellcolor{gray!10}43.45 & 41.39 & 54.92$\ddel{55.07}$ & \cellcolor{blue!4}\textbf{58.71}$\ddel{58.12}$ & \cellcolor{gray!10}41.83& 40.16  & 54.13$\ddel{54.33}$ & \cellcolor{blue!4}55.17$\dbb{55.76}$ & \cellcolor{gray!10}29.04 & 28.54 & 48.67$\ddel{49.02}$ & \cellcolor{blue!4}49.85$\dbb{50.44}$ & \cellcolor{gray!10}65.01 & 59.30 & 66.83$\ddel{68.21}$ & \cellcolor{blue!4}66.88$\dbb{68.95}$\\
        % & Top-10 & \cellcolor{gray!10}0.86 & 0.86 & 0.86 & \cellcolor{gray!10}0.82 & 0.77 & 0.83 & \cellcolor{gray!10}0.91 & 0.88 & 0.84 & \cellcolor{gray!10}0.78 & 0.71 & 0.82 \\
        & Top-20 & \cellcolor{gray!10}65.70 & 65.50 & 72.69$\ddel{72.44}$ & \cellcolor{blue!4}\textbf{73.43}$\ddel{72.79}$ & \cellcolor{gray!10}62.40 & 62.35 & 68.50$\ddel{68.55}$ & \cellcolor{blue!4}\textbf{69.54}$\ddel{69.14}$ & \cellcolor{gray!10}49.21 & 49.51 & 63.19$\ddel{63.24}$ & \cellcolor{blue!4}\textbf{62.40}$\dbb{62.40}$ & \cellcolor{gray!10}74.61 & 74.46 & 76.67$\ddel{76.53}$ & \cellcolor{blue!4}76.96$\dbb{77.36}$ \\
        % & Top-50 & \cellcolor{gray!10}0.86 & 0.86 & 0.86 & \cellcolor{gray!10}0.82 & 0.77 & 0.83 & \cellcolor{gray!10}0.91 & 0.88 & 0.84 & \cellcolor{gray!10}0.78 & 0.71 & 0.82 \\
        \cmidrule(lr){2-18}
        & nDCG@1 & \cellcolor{gray!10}19.98 & 18.65 & 26.62$\ddel{28.05}$ & \cellcolor{blue!4}\textbf{32.53}$\ddel{30.81}$ & \cellcolor{gray!10}18.90 & 17.32 & 27.56$\ddel{28.54}$ & \cellcolor{blue!4}\textbf{33.91}$\ddel{33.56}$ & \cellcolor{gray!10}11.66 & 11.96 & 26.38$\ddel{25.44}$ & \cellcolor{blue!4}\textbf{29.38}$\ddel{27.66}$ & \cellcolor{gray!10}\textcolor{red}{44.83} & 37.16 & 39.32$\ddel{46.26}$ & \cellcolor{blue!4}42.18$\dbb{48.03}$ \\
        % & nDCG@3 & \cellcolor{gray!10}0.86 & 0.86 & 0.86 & \cellcolor{gray!10}0.82 & 0.77 & 0.83 & \cellcolor{gray!10}0.91 & 0.88 & 0.84 & \cellcolor{gray!10}0.78 & 0.71 & 0.82 \\
        & nDCG@5 & \cellcolor{gray!10}18.64 & 17.44 & 26.78$\ddel{26.90}$ & \cellcolor{blue!4}\textbf{30.82}$\ddel{29.89}$ & \cellcolor{gray!10}19.36 & 17.95 & 27.39$\ddel{28.27}$ & \cellcolor{blue!4}30.72$\dbb{30.86}$ & \cellcolor{gray!10}11.57 & 10.81 & 26.67$\ddel{26.08}$ & \cellcolor{blue!4}\textbf{28.21}$\ddel{27.53}$ & \cellcolor{gray!10}39.76 & 34.35 & 38.66$\ddel{42.59}$ & \cellcolor{blue!4}40.34$\dbb{43.67}$ \\
        % & nDCG@10 & \cellcolor{gray!10}0.86 & 0.86 & 0.86 & \cellcolor{gray!10}0.82 & 0.77 & 0.83 & \cellcolor{gray!10}0.91 & 0.88 & 0.84 & \cellcolor{gray!10}0.78 & 0.71 & 0.82 \\
        & nDCG@20 & \cellcolor{gray!10}22.22 & 21.53 & 31.18$\ddel{31.06}$ & \cellcolor{blue!4}\textbf{33.79}$\ddel{33.21}$ & \cellcolor{gray!10}22.12 & 21.41 & 31.44$\ddel{31.82}$ & \cellcolor{blue!4}\textbf{33.62}$\ddel{33.43}$ & \cellcolor{gray!10}14.84 & 14.45 & 32.46$\ddel{31.83}$ & \cellcolor{blue!4}\textbf{33.20}$\ddel{32.45}$ & \cellcolor{gray!10}38.95 & 36.21 & 41.81$\ddel{44.32}$ & \cellcolor{blue!4}42.65$\dbb{44.74}$ \\
        % & nDCG@50 & \cellcolor{gray!10}0.86 & 0.86 & 0.86 & \cellcolor{gray!10}0.82 & 0.77 & 0.83 & \cellcolor{gray!10}0.91 & 0.88 & 0.84 & \cellcolor{gray!10}0.78 & 0.71 & 0.82 \\
        % & nDCG@100 & \cellcolor{gray!10}0.86 & 0.86 & 0.86 & \cellcolor{gray!10}0.82 & 0.77 & 0.83 & \cellcolor{gray!10}0.91 & 0.88 & 0.84 & \cellcolor{gray!10}0.78 & 0.71 & 0.82 \\
        \cmidrule(lr){2-18}
        & MAP@100 & \cellcolor{gray!10}18.79 & 18.22 & 25.92$\ddel{25.62}$ & \cellcolor{blue!4}\textbf{27.82}$\ddel{27.24}$ & \cellcolor{gray!10}19.15 & 18.39 & 26.63$\ddel{26.81}$ & \cellcolor{blue!4}\textbf{28.09}$\ddel{28.02}$ & \cellcolor{gray!10}12.03& 11.53  & 26.20$\ddel{25.32}$ & \cellcolor{blue!4}\textbf{26.84}$\ddel{25.95}$ & \cellcolor{gray!10}33.32 & 30.44 & 36.46$\ddel{38.58}$ & \cellcolor{blue!4}36.82$\dbb{38.66}$ \\
        \toprule[0.5pt]
        \multirow{7}{*}{TriviaQA}& Top-1 & \cellcolor{gray!10}34.16 & 25.17 & 51.77$\ddel{51.17}$ & \cellcolor{blue!4}\textbf{54.95}$\ddel{53.99}$ & \cellcolor{gray!10}46.30 & 35.10 & 55.85$\ddel{57.76}$ & \cellcolor{blue!4}58.70$\dbb{59.80}$ & \cellcolor{gray!10}30.76 & 21.19 & 52.84$\ddel{52.74}$ & \cellcolor{blue!4}\textbf{54.35}$\ddel{53.83}$ & \cellcolor{gray!10}57.47 & 37.16 & 62.55$\ddel{66.77}$ & \cellcolor{blue!4}63.47$\dbb{67.23}$  \\
        % & Top-3 & \cellcolor{gray!10}0.86 & 0.86 & 0.86 & \cellcolor{gray!10}0.82 & 0.77 & 0.83 & \cellcolor{gray!10}0.91 & 0.88 & 0.84 & \cellcolor{gray!10}0.78 & 0.71 & 0.82 \\
        & Top-5 & \cellcolor{gray!10}59.49 & 50.99 & 73.81$\ddel{73.69}$ & \cellcolor{blue!4}\textbf{74.31}$\ddel{74.02}$ & \cellcolor{gray!10}66.28 & 57.64 & 75.60$\ddel{75.98}$ & \cellcolor{blue!4}\textbf{76.04}$\ddel{75.86}$ & \cellcolor{gray!10}52.65 & 43.16 & 70.94$\ddel{70.78}$ & \cellcolor{blue!4}\textbf{71.12}$\ddel{70.78}$ & \cellcolor{gray!10}72.40 & 58.84 & 78.74$\ddel{79.06}$ & \cellcolor{blue!4}78.84$\dbb{79.19}$ \\
        % & Top-10 & \cellcolor{gray!10}0.86 & 0.86 & 0.86 & \cellcolor{gray!10}0.82 & 0.77 & 0.83 & \cellcolor{gray!10}0.91 & 0.88 & 0.84 & \cellcolor{gray!10}0.78 & 0.71 & 0.82 \\
        & Top-20 & \cellcolor{gray!10}73.91 & 74.10 & 80.08$\ddel{80.01}$ & \cellcolor{blue!4}\textbf{80.22}$\ddel{80.16}$ & \cellcolor{gray!10}76.41 & 76.24 & 80.68$\dbb{80.70}$ & \cellcolor{blue!4}80.66$\dbb{80.70}$ & \cellcolor{gray!10}67.18 & 67.44 & 76.34$\ddel{76.28}$ & \cellcolor{blue!4}\textbf{76.32}$\ddel{76.27}$ & \cellcolor{gray!10}79.77 & 79.66 & 83.13$\ddel{83.07}$ & \cellcolor{blue!4}83.15$\dbb{83.16}$\\
        % & Top-50 & \cellcolor{gray!10}0.86 & 0.86 & 0.86 & \cellcolor{gray!10}0.82 & 0.77 & 0.83 & \cellcolor{gray!10}0.91 & 0.88 & 0.84 & \cellcolor{gray!10}0.78 & 0.71 & 0.82 \\
        \cmidrule(lr){2-18}
        & nDCG@1 & \cellcolor{gray!10}34.16 & 25.17 & 51.77$\ddel{51.17}$ & \cellcolor{blue!4}\textbf{54.95}$\ddel{53.99}$ & \cellcolor{gray!10}46.30 & 35.10 & 55.85$\ddel{57.76}$ & \cellcolor{blue!4}58.70$\dbb{59.80}$ & \cellcolor{gray!10}30.76 & 21.19 & 52.84$\ddel{52.74}$ & \cellcolor{blue!4}\textbf{54.35}$\ddel{53.83}$ & \cellcolor{gray!10}57.74 & 37.16 & 62.55$\ddel{66.77}$ & \cellcolor{blue!4}63.47$\dbb{67.23}$ \\
        % & nDCG@3 & \cellcolor{gray!10}0.86 & 0.86 & 0.86 & \cellcolor{gray!10}0.82 & 0.77 & 0.83 & \cellcolor{gray!10}0.91 & 0.88 & 0.84 & \cellcolor{gray!10}0.78 & 0.71 & 0.82 \\
        & nDCG@5 & \cellcolor{gray!10}30.46 & 23.63 & 49.27$\ddel{48.52}$ & \cellcolor{blue!4}\textbf{51.21}$\ddel{50.23}$ & \cellcolor{gray!10}41.60 & 32.77 & 53.55$\ddel{56.65}$ & \cellcolor{blue!4}55.28$\dbb{55.89}$ & \cellcolor{gray!10}27.78 & 19.76 & 50.47$\ddel{50.42}$ & \cellcolor{blue!4}\textbf{51.60}$\ddel{51.09}$ & \cellcolor{gray!10}49.69 & 34.42 & 59.53$\ddel{61.93}$ & \cellcolor{blue!4}59.99$\dbb{62.16}$ \\
        % & nDCG@10 & \cellcolor{gray!10}0.86 & 0.86 & 0.86 & \cellcolor{gray!10}0.82 & 0.77 & 0.83 & \cellcolor{gray!10}0.91 & 0.88 & 0.84 & \cellcolor{gray!10}0.78 & 0.71 & 0.82 \\
        & nDCG@20 & \cellcolor{gray!10}31.78 & 28.69 & 50.92$\ddel{50.36}$ & \cellcolor{blue!4}\textbf{52.06}$\ddel{51.43}$ & \cellcolor{gray!10}40.68 & 36.65 & 54.93$\ddel{55.53}$ & \cellcolor{blue!4}55.66$\dbb{55.93}$ & \cellcolor{gray!10}29.25 & 25.80 & 53.12$\ddel{52.80}$ & \cellcolor{blue!4}\textbf{53.62}$\ddel{53.12}$ & \cellcolor{gray!10}46.33 & 39.64 & 59.90$\ddel{61.14}$ & \cellcolor{blue!4}60.06$\dbb{61.18}$ \\
        % & nDCG@50 & \cellcolor{gray!10}0.86 & 0.86 & 0.86 & \cellcolor{gray!10}0.82 & 0.77 & 0.83 & \cellcolor{gray!10}0.91 & 0.88 & 0.84 & \cellcolor{gray!10}0.78 & 0.71 & 0.82 \\
        % & nDCG@100 & \cellcolor{gray!10}0.86 & 0.86 & 0.86 & \cellcolor{gray!10}0.82 & 0.77 & 0.83 & \cellcolor{gray!10}0.91 & 0.88 & 0.84 & \cellcolor{gray!10}0.78 & 0.71 & 0.82 \\
        \cmidrule(lr){2-18}
        & MAP@100 & \cellcolor{gray!10}26.61 & 23.85 & 44.69$\ddel{43.93}$ & \cellcolor{blue!4}\textbf{45.58}$\ddel{44.81}$ & \cellcolor{gray!10}34.85 & 30.98 & 49.50$\ddel{49.91}$ & \cellcolor{blue!4}49.93$\dbb{50.11}$ & \cellcolor{gray!10}24.02 & 20.83 & 47.02$\ddel{46.47}$ & \cellcolor{blue!4}\textbf{47.45}$\ddel{46.71}$ & \cellcolor{gray!10}39.40 & 33.01 & 54.21$\ddel{55.34}$ & \cellcolor{blue!4}54.31$\dbb{55.42}$ \\
        \bottomrule[1pt]
    \end{tabular}
     \begin{tablenotes}
        \footnotesize
        \item[*] Orig. indicates the original results from the retriever, where no re-ranking method is employed.
    \end{tablenotes}
     \end{threeparttable}
    }
    \vspace{-3mm}
    \caption{Re-ranking results on the test set of datasets of the Top-100 retrieved documents with the LLaMA2-7B model. The best results are highlighted in bold. The higher scores of original retriever compared with $\mathrm{UR^3}$ is highlighted in red. The results on other models (Mistral-7B and GPT-Neo) are detailed in Appendix~\ref{appendix:reranking}.}
    \label{tab:main-results}
    \vspace{-3mm}
\end{table*}
\addtolength{\tabcolsep}{0.3em}
% \vspace{10em}

\begin{figure*}[t]
    \centering
    \begin{subfigure}[b]{0.32\textwidth}
        \includegraphics[width=\textwidth]{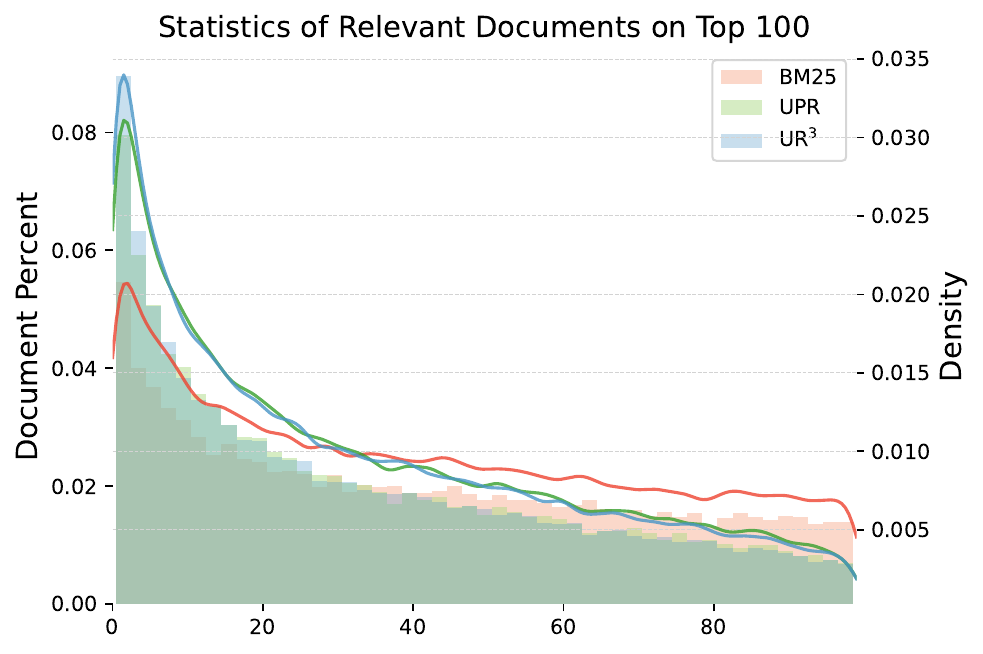}
        \caption{Relevant Document Distribution}
        \label{fig:dist}
    \end{subfigure}
    \hfill % 添加适当的空格
    \begin{subfigure}[b]{0.32\textwidth}
        \includegraphics[width=\textwidth]{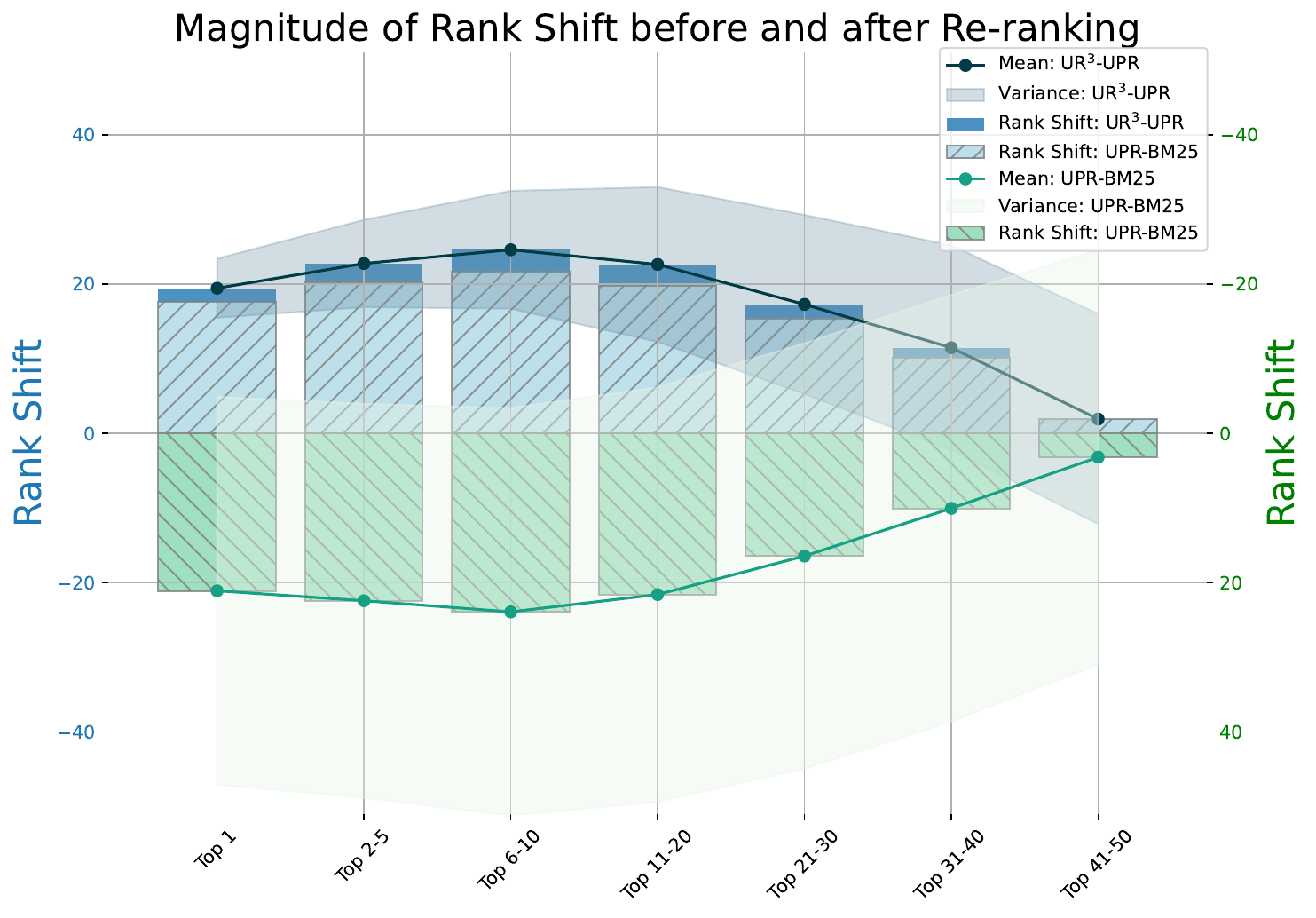}
        \caption{Rank Shift of Relevant Document}
        \label{fig:shift}
    \end{subfigure}
    \hfill % 添加适当的空格
    \begin{subfigure}[b]{0.32\textwidth}
        \includegraphics[width=\textwidth]{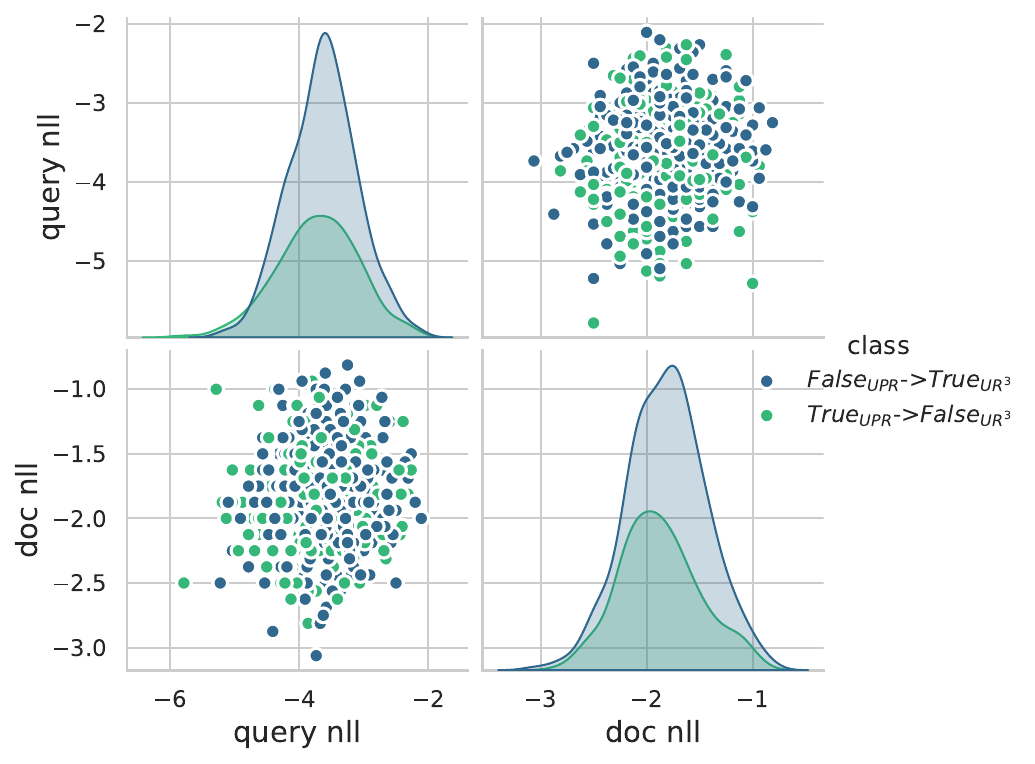}
        \caption{Document Distribution on Top-1}
        \label{fig:scatt}
    \end{subfigure}
    \caption{Visualization of Analysis on the Enhanced Performance in the Re-ranking task}
\end{figure*}

\subsection{Document Re-ranking}\label{exp:rerank}
We evaluate the performance of our $\mathrm{UR^3}$ method across all evaluated datasets and retrievers.
\subsubsection{Overall Performance}
\paragraph{Comprehensive better than UPR.}
As shown in the Table~\ref{tab:main-results}, the results demonstrate that $\mathrm{UR^3}$ enhances the overall rankings of the Top-100 documents, as reflected by an average increase of 1-2\% in the MAP@100 metric.
Furthermore, improvements are observed across all nDCG@K metrics, indicating that $\mathrm{UR^3}$ prioritizes relevant documents more effectively compared to the UPR method. Closer examination of the Top-K metrics reveals that $\mathrm{UR^3}$ shows greater accuracy enhancements for rankings closer to the top, with the most substantial increase (up to 6.64) observed at Top-1 accuracy. This significantly enhances the suitability of our method for open-domain question answering tasks. Additionally, it potentially alleviates the issues associated with the limited input window length of large models, as our method achieves higher relevance scores with fewer input documents.

\paragraph{Why does RankGPT perform poorly?}
Interestingly, the RankGPT method yields lower ranking results than the initial retrieval. This can be attributed to the observation that competitive performance is predominantly realized by model based on GPT-4~\cite{journals/corr/abs-2308-07107}. When utilizing smaller parameterized language models, such as LLaMA2-7B, the RankGPT method underperforms compared to other methods.

\paragraph{Unstable performance on Interpolation.}
The degree of increase (or decrease) brought by the Interpolation method primarily depends on the performance of the initial retriever. For the supervised DPR retriever, its exposure to relevant paragraphs during training yields substantially higher Top-1 accuracy on the NQ and WebQ datasets. With results from the DPR, the Interpolation method usually brings a significant enhancement in ranking. When combined with our $\mathrm{UR^3}$ method, this can lead to maximal improvement. Conversely, when based on an unsupervised retriever, our method predominantly outperforms the Interpolation method.
% which are consistently observed across both Unsupervised Ranking Refinement (UPR) and $\mathrm{UR^3}$.
% conversely, unsupervised retrievers like Contriever and MSS may lead to negative gains. 

% Compared to both the method without re-ranking and the baseline re-ranking method, the proposed method consistently achieves higher accuracy and better ranking quality, as evidenced by improvements in nDCG and MAP scores.

\begin{table*}[ht]
\addtolength{\tabcolsep}{-0.45pt}
\tiny
\centering
\resizebox{\textwidth}{!}{%
\begin{tabular}{lcccccc|cccccc|cccccc}
 \toprule
 \textbf{Model} & \multicolumn{6}{c}{\textbf{LLaMA2-13B}} & \multicolumn{6}{c}{\textbf{Mistral-7B}} & \multicolumn{6}{c}{\textbf{Gemma-7B}}\\
    \cmidrule(lr){2-7} \cmidrule(lr){8-13} \cmidrule(lr){14-19} 
   & \multicolumn{2}{c}{\textbf{NQ}} & \multicolumn{2}{c}{\textbf{WebQ}} & \multicolumn{2}{c}{\textbf{TriviaQA}} & \multicolumn{2}{c}{\textbf{NQ}} & \multicolumn{2}{c}{\textbf{WebQ}} & \multicolumn{2}{c}{\textbf{TriviaQA}} & \multicolumn{2}{c}{\textbf{NQ}} & \multicolumn{2}{c}{\textbf{WebQ}} & \multicolumn{2}{c}{\textbf{TriviaQA}} \\
    &  EM & F1 & EM & F1 & EM & F1 &  EM & F1 & EM & F1 & EM & F1 &  EM & F1 & EM & F1 & EM & F1\\
\midrule
\rowcolor{gray!10}Contriever & 22.02 & 29.11  &  19.69 & 30.21 & 49.90  &  57.08 & 20.69 & 26.61 & 14.37 & 24.39 &49.89 & 56.94 & 17.40 & 25.13 & 14.71 & 26.05 & 45.54 & 53.66\\
\phantom{0000}+ Inference with UPR         & 28.06 & 35.99  &  22.00 & 33.48 & 58.89 & 67.05 & 24.68 & 31.81 & 15.90 & 25.80 &59.20 &66.83 & 21.33 &29.72 & 15.29 & 27.05 & 55.26 &64.05\\
\phantom{0000}+ Inference with $\mathrm{UR^3}$         & \textbf{28.45} & \textbf{37.05}  &  \textbf{23.18} & \textbf{34.92} & \textbf{59.45} & \textbf{67.72} & \textbf{25.51} & \textbf{32.69} & \textbf{17.13} & \textbf{27.10} & \textbf{59.26} & \textbf{67.09} & \textbf{22.13} & \textbf{30.78} & \textbf{15.99} & \textbf{27.65} & \textbf{55.47} & \textbf{64.53}\\
\midrule
\rowcolor{gray!10}BM25 & 20.20 & 27.08  &  16.39 & 26.60 & 55.21  &  62.91 & 19.14 & 25.31 & 13.29 & 23.03 & 54.91 & 62.15 &16.40 & 23.84 & 12.35 & 22.84 & 52.34 & 60.89\\
\phantom{0000}+ Inference with UPR         & 27.23 & 34.92  &  19.98 & 31.27 &61.86 &69.96 & 24.79 & 31.71 & 15.21 & 25.43 & 61.68 & 69.16 & 21.02 & 29.47 & 14.67 & 25.32 & 58.53 & 67.36\\
\phantom{0000}+ Inference with $\mathrm{UR^3}$         & \textbf{28.37} & \textbf{36.69}  &  \textbf{21.46} & \textbf{32.83} & \textbf{62.34} & \textbf{70.62} & \textbf{25.73} & \textbf{32.87} & \textbf{16.49} & \textbf{26.69} & \textbf{61.81} & \textbf{69.50} & \textbf{22.30} & \textbf{30.96} & \textbf{15.40} & \textbf{25.80} & \textbf{58.99} & \textbf{67.83}\\
\midrule
\rowcolor{gray!10}MSS & 19.86 & 26.16  &  16.83  & 27.94 & 49.29  &  56.03 & 18.20 & 24.28 & 13.98 & 23.79 & 48.41 & 55.22 & 14.38 & 21.44 & 12.20 & 22.91 & 43.56 & 51.54\\
\phantom{0000}+ Inference with UPR         & 26.81 & 34.63  &  20.37 & 31.77 &57.87 &65.69 & 23.35 & 30.04 & 16.04 & 26.24 & 57.46 & 65.11 & 19.83 & 28.65 & 14.16 & 26.20 & 53.84 & 62.67\\
\phantom{0000}+ Inference with $\mathrm{UR^3}$         & \textbf{27.26} & \textbf{35.35}  &  \textbf{21.16} & \textbf{32.81} & \textbf{58.28} & \textbf{66.15} & \textbf{24.29} & \textbf{31.22} & \textbf{16.58} & \textbf{26.72} & \textbf{57.69} & \textbf{65.34} & \textbf{21.27} & \textbf{29.69} & \textbf{15.06} & \textbf{26.29} & \textbf{53.97} & \textbf{62.77}\\
\midrule
\rowcolor{gray!10}DPR & 30.30 & 38.42  &  22.79  & 34.36 & 55.33  &  62.96 & \textbf{\textcolor{red}{28.17}} & 34.9 & 18.21 & \textbf{\textcolor{red}{28.55}} & 55.03 & 62.24 & \textbf{\textcolor{red}{24.43}} & \textbf{\textcolor{red}{33.14}} & \textbf{\textcolor{red}{16.68}} & 28.00 & 50.76 & 59.59\\
\phantom{0000}+ Inference with UPR         & 29.09 & 37.21  &  23.18 &34.09 &60.87 &69.02 & 26.51 & 34.13 & 16.98 & 26.78 & 60.95 & 68.71 & 22.33 & 31.36 & 15.99 & 27.44 & 57.23 & 66.19\\
\phantom{0000}+ Inference with $\mathrm{UR^3}$         & \textbf{30.80} & \textbf{39.23}  &  \textbf{24.36} & \textbf{36.15} & \textbf{61.21} & \textbf{69.35} & 27.84 & \textbf{35.36} & \textbf{18.31} & 28.13 & \textbf{61.12} & \textbf{68.94} & 23.91 & 32.80 & 16.54 & \textbf{28.16} & \textbf{57.43} & \textbf{66.40}\\
 \bottomrule
\end{tabular}
}
\caption{EM and F1 scores for the open-domain QA task. We perform inference with the re-ranked Top-1 results of Table~\ref{tab:main-results}. The best performing models are highlighted in bold. We highlight the best scores obtained by original retriever in red. 
% Furthermore, we explore different number of input documents with LLaMA2-7B model in Table~\ref{tab:odqa-llama2-7b}. 
We also conduct inference on the re-ranking results of Mistral-7B in Table~\ref{tab:odqa-mistral}.}
\label{tab:odqa-ansgen}
%\vspace{-6pt}
\end{table*}

\begin{table} [t]
% \addtolength{\tabcolsep}{-0.45pt}
\tiny
\centering
\resizebox{\columnwidth}{!}{%
\begin{tabular}{lcccccc}
 \toprule
 % \textbf{Model} & \multicolumn{6}{c}{\textbf{LLaMA2-B}} & \multicolumn{6}{c}{\textbf{Mistral-7B}} & \multicolumn{6}{c}{\textbf{Gemma-7B}}\\
    % \cmidrule(lr){2-7} \cmidrule(lr){8-13} \cmidrule(lr){14-19} 
  \textbf{NQ Dataset} & \multicolumn{2}{c}{\textbf{Top-1}} & \multicolumn{2}{c}{\textbf{Top-3}} & \multicolumn{2}{c}{\textbf{Top-5}}  \\
  \cmidrule(lr){2-3} \cmidrule(lr){4-5} \cmidrule(lr){6-7} 
    &  EM & F1 & EM & F1 & EM & F1 \\
% \midrule
\midrule
\rowcolor{gray!10}Contriever & 15.09 & 22.00  &  14.93 & 20.36 & 18.50  &  23.80 \\
\phantom{0000}+ Inference with UPR         & 20.97 & 27.90  &  19.31 & 25.51 & 22.33 & 28.61 \\
\phantom{0000}+ Inference with $\mathrm{UR^3}$         & \textbf{21.93} & \textbf{29.06}  &  \textbf{19.98} & \textbf{25.77} & \textbf{22.55} & \textbf{28.72} \\
\midrule
\rowcolor{gray!10}BM25 & 15.65 & 21.38  &  14.35 & 20.04 & 16.45  &  22.05\\
\phantom{0000}+ Inference with UPR         & 20.75 & 27.71  &  19.56 & 25.90 & 21.91 & 28.28\\
\phantom{0000}+ Inference with $\mathrm{UR^3}$         & \textbf{21.75} & \textbf{28.91}  &  \textbf{20.02} & \textbf{26.90} & \textbf{22.27} & \textbf{28.70} \\
\midrule
\rowcolor{gray!10}MSS & 13.60 & 19.34  &  13.63  & 19.48 & 16.59  &  22.22\\
\phantom{0000}+ Inference with UPR         & 19.78 & 26.98  &  18.70 & 24.96 & 20.72 & 26.87\\
\phantom{0000}+ Inference with $\mathrm{UR^3}$         & \textbf{21.69} & \textbf{28.66}  &  \textbf{19.47} & \textbf{26.20} & \textbf{21.27} & \textbf{27.68} \\
\midrule
\rowcolor{gray!10}DPR & 23.38 & 30.69  &  19.61  & 26.21 & 22.60  &  28.84 \\
\phantom{0000}+ Inference with UPR         & 22.13 & 29.58  &  20.42 & 27.19 & 23.74 & 30.21 \\
\phantom{0000}+ Inference with $\mathrm{UR^3}$         & \textbf{24.29} & \textbf{31.16}  &  \textbf{22.08} & \textbf{29.24} & \textbf{24.54} & \textbf{30.96} \\
 \bottomrule
\end{tabular}
}
\caption{EM and F1 scores for the open-domain QA task with different number of input documents on the NQ dataset with LLaMA2-7B model. The best performing models are highlighted in bold.}
\label{tab:odqa-llama2-7b-nq}
%\vspace{-6pt}
\end{table}

\subsubsection{Analysis on the Enhanced Performance}
To further explore why $\mathrm{UR^3}$ demonstrates greater enhancements for the ranks close to the top, we conduct empirical analyses on the NQ dataset with BM25 retriever. 

As illustrated in Figure~\ref{fig:dist}, we analyze the distributional shift of relevant document positions before and after re-ranking. The histogram represents the proportion of relevant documents at different ranks, and a curve fitting illustrates the trend of this distributional change. Overall, it is evident that $\mathrm{UR^3}$ tends to shift the distribution of relevant documents towards higher ranks compared to UPR. 

Then we explore the reason behind the forward shift in this distribution. In Figure~\ref{fig:shift}, we quantify the magnitude of rank shifts for each relevant document after re-ranking. The blue solid (shadowed) histogram represents the positional change in rank by $\mathrm{UR^3}$ compared to UPR (BM25), while the green shadowed histogram indicates the change by UPR compared to BM25. The bandwidth of the line graph represents the variance of these changes. The figure clearly shows that $\mathrm{UR^3}$ induces smaller shifts in each position; in other words, our method tunes the rankings of relevant documents within a narrower range, thereby obtaining greater benefits to the ranks closer to the top (as the distribution of relevant documents is higher in Figure~\ref{fig:dist}).

Figure~\ref{fig:scatt} presents a scatter plot that statistically categorizes the relevant documents at the Top-1 rank, comparing UPR and $\mathrm{UR^3}$. Each green dot represents a correct calibration by the $\mathrm{UR^3}$ method, where an irrelevant document ranked by UPR is adjusted to a relevant one at the Top-1 rank. Conversely, each blue dot indicates a incorrect calibration by $\mathrm{UR^3}$, where a previously Top-1 relevant document is replaced with an irrelevant one. The axes values denote the respective query/document generation negative log-likelihood loss (nll) discussed in Formula~\ref{eq:total}. The density distribution of the scatter plot reveals that the positive gains brought about by $\mathrm{UR^3}$ significantly outweigh the negative impacts, which substantiates the improvement observed at the Top-1.

\subsection{Application in Question Answering}\label{exp:qa}
As discussed above, we have demonstrated that $\mathrm{UR^3}$ significantly enhances ranking performance. In this section, we apply the results of the re-ranking (Table~\ref{tab:main-results}) to apply in current RAG models for the evaluation in open-domain QA tasks. 
Specifically, we utilize the Top-n (n $\leq$ 5) items as inputs to achieve higher scores with fewer documents. 

\subsubsection{Overall Performance}\label{sec:qa-performance}
\paragraph{Not More is Better.} We utilize different number of document inputs on NQ dataset to evaluate the QA performance in Table~\ref{tab:odqa-llama2-7b-nq}. Expanding the documents from Top-1 to Top-3 does not invariably enhance performance; in fact, it occasionally results in a decline in both EM and F1 scores. This trend suggests that increasing the number of documents beyond the most relevant one may introduce noise or less pertinent information. Furthermore, the marginal gains observed when moving from Top-1 to Top-5 are minimal, which underscores the diminishing returns of adding more documents. In summary, utilizing the Top-1 document emerges as the most cost-effective approach, offering a balance between computational efficiency and accuracy.
\paragraph{Superiority Over UPR.}
% We conduct experiments using re-ranked results of Table~\ref{tab:main-results}.
As illustrated in the Table~\ref{tab:odqa-ansgen} and \ref{tab:odqa-llama2-7b-nq}, the $\mathrm{UR^3}$ method substantially enhances the performance of QA tasks, achieving superior EM and F1 scores compared to the UPR method.  Furthermore, this improvement trend is consistent with the enhancements observed during the re-ranking phase.
\paragraph{Outliers on DPR.}
In Table~\ref{tab:main-results}, it is noteworthy that the highest scores (indicated in red) are achieved by the original DPR method on the Mistral and Gemma models. This is explainable given that both UPR and $\mathrm{UR^3}$ exhibit inferior re-ranking performance compared to DPR for the Top-1 results. However, when employing the LLaMA2-13B model, it demonstrates superior QA performance relative to the DPR. This improvement can be attributed to the strategic use of a generative model with a distribution similar to that of the re-ranker (e.g., within the same LLaMA2 series) in a QA task. In $\mathrm{UR^3}$, maximizing the probability of document generation has a benefit to selecting documents that closely align with the model’s distribution. Such alignment significantly enhances the model’s reliance on external documents, thereby boosting the overall performance of the QA task. 
% Further demonstration of the analysis are detailed in Section~\ref{exp:analysis}.

\begin{figure}
    \centering
    \begin{minipage}{0.46\textwidth}
        \includegraphics[width=\textwidth]{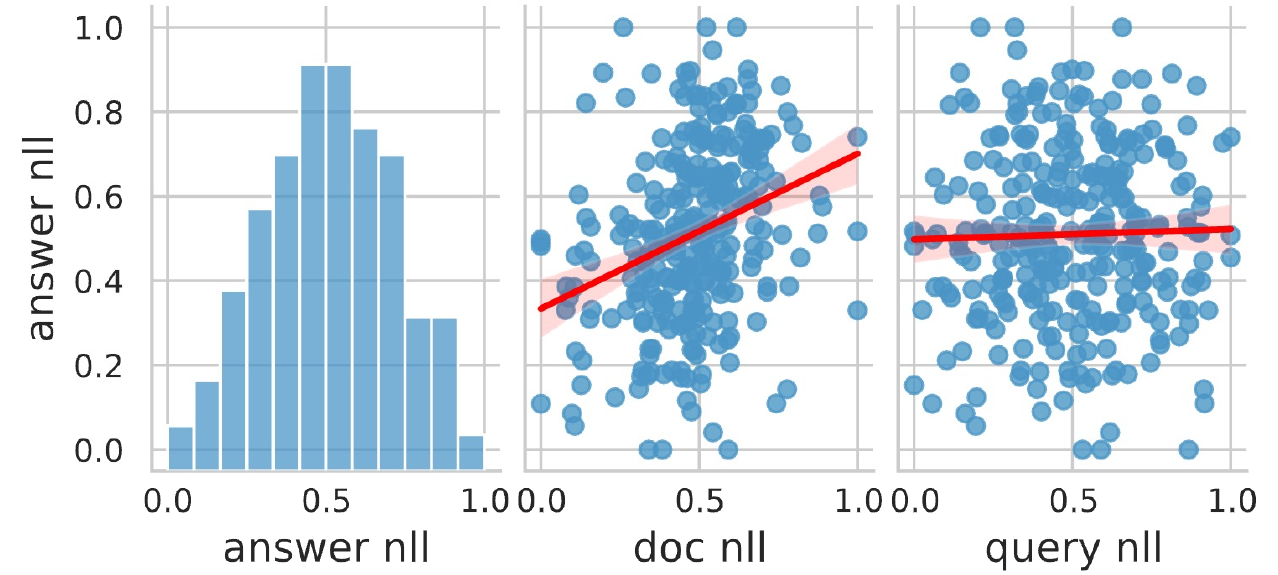}
        \caption{Distributed correlation in answer generation with normalized NLL in the QA task.}
        \label{fig:qa_scatt}
    \end{minipage}
\end{figure}

% \subsection{Analyses on the Enhanced Performance}

% To further explore why $\mathrm{UR^3}$ demonstrates greater enhancements for re-ranking and QA tasks, we conduct empirical analyses on the NQ dataset with BM25. 

\subsubsection{Analysis on the DPR results}\label{exp:analysis}
We conduct empirical analysis for the improved performance on DPR retriever with LLaMA2-13B. Figure~\ref{fig:qa_scatt} presents the distributed correlation in answer generation with normalized negative log-likelihood loss of the QA task. When the generation probability is high, the corresponding loss is low. 
The left panel displays the distribution of answer NLL values. The middle and right panels feature scatter plots that illustrate the relationships between document generation NLL (doc NLL) and query generation NLL (query NLL) during the re-ranking phase. Both scatter plots include a regression line, indicating that, compared to query loss, document loss shows a positive correlation. This suggests that higher generation probabilities for documents increase the likelihood of generating the correct answer. This finding aligns with our understanding that selecting documents closely matching the model’s distribution can enhance the model’s receptivity to external documents.
% The regression lines in these plots serve as visual confirmation of these trends, which are essential for diagnosing and enhancing model performance in the QA task.

\subsection{Accuracy and Efficiency Comparison}\label{sec:efficiency}
\begin{figure}[t] 
\centering
  \includegraphics[width=0.9\columnwidth]{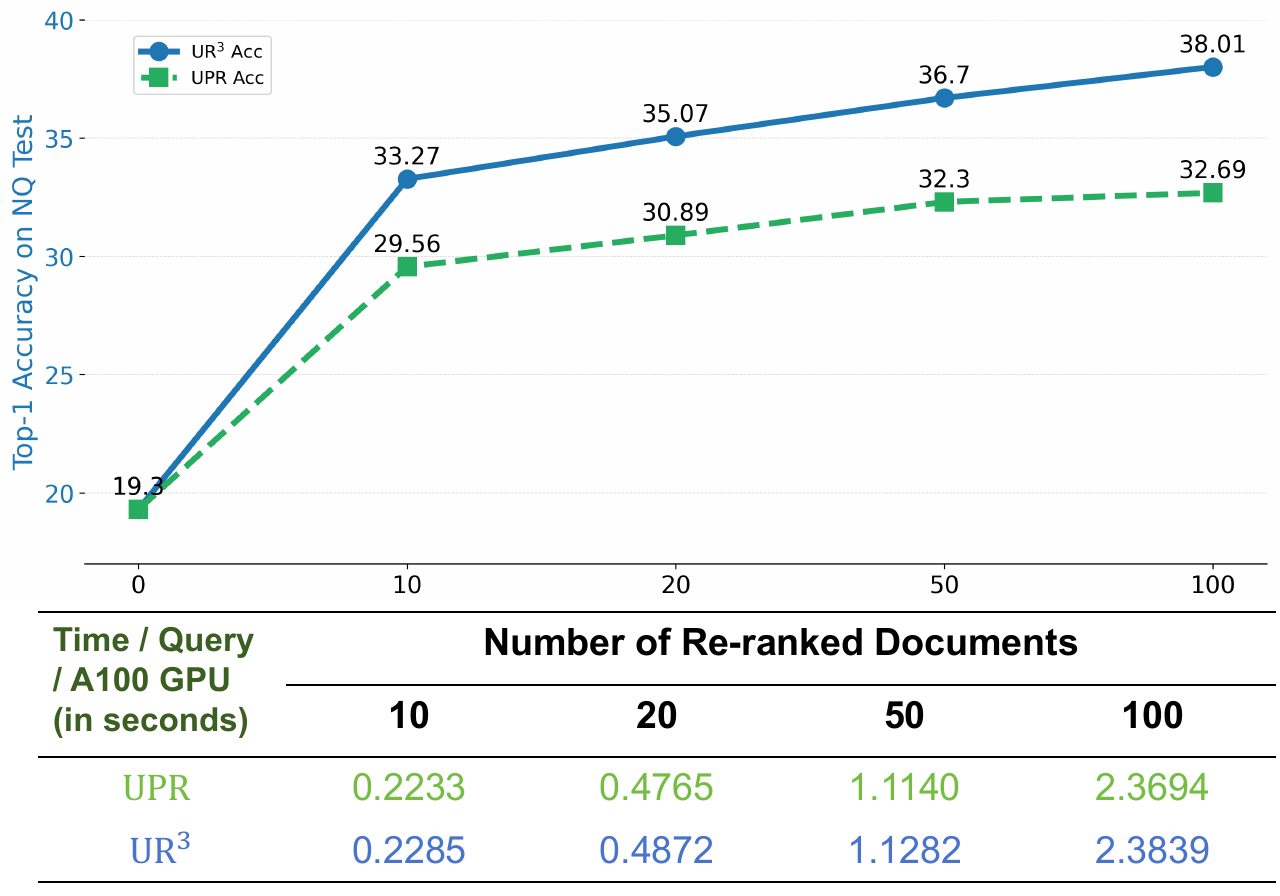}
  \caption{Effect of the number of document candidates on Top-1 accuracy and calculation efficiency when re-ranked with LLaMA2-7B model. Evaluation is done on the NQ test set using BM25 retrieved documents.}
  \label{fig:time}
\end{figure}
In this section, we evaluate the impact of the number of document candidates to be re-ranked on both retrieval performance and computational efficiency. The evaluation is conducted using the NQ test set. We re-rank up to the Top-100 documents obtained from the BM25 retriever and measure performance using Top-1 accuracy.

In Figure~\ref{fig:time}, as the number of re-ranked documents increases, both $\mathrm{UR^3}$ and UPR exhibit improvements in Top-1 accuracy.
$\mathrm{UR^3}$ consistently outperforms UPR across all document counts, achieving higher Top-1 accuracy.
On the other hand, the computational time per query increases linearly with the number of re-ranked documents for both methods.
Despite the increase in computational time, $\mathrm{UR^3}$ maintains a similar computational demand compared to UPR.

In conclusion, the results clearly show that $\mathrm{UR^3}$ significantly enhances performance without incurring additional computational time, which shows the superiority of our method.

\section{Conclusion}
In this study, we introduced the $\mathrm{UR^3}$ framework, which utilizes Bayesian decision theory to address the estimation bias in QLMs based on LLMs. The novelty of $\mathrm{UR^3}$ lies in its approach to unify the probabilities of query and document generation under a risk minimization framework, thereby enhancing the efficiency of document ranking and question answering. 

Our experimental results demonstrate that $\mathrm{UR^3}$ significantly improves re-ranking performance, especially in terms of Top-1 accuracy. In open-domain question-answering tasks, $\mathrm{UR^3}$ contributes to achieving higher accuracy with reduced reliance on the number of input documents.

\section*{Limitations}
\begin{itemize}
    \item This paper observes relatively minor improvements when ranking is extended to Top-20 or Top-50, marking a principal limitation. However, a longer context does not necessarily equate to superior performance for the LLMs~\cite{liu2024lost}, which have also been discussed in Section~\ref{sec:qa-performance}. Our method achieves a substantial improvement in Top-1 accuracy with comprehensive analysis, which provides optimal cost-effectiveness.
    \item A limitation of $\mathrm{UR^3}$ is that re-ranking a large pool of document can have a high latency as it involves performing cross-attention whose complexity is proportional to the product of the question and document tokens and the number of layers of the LLM. We have also discussed this quantitatively in Section~\ref{sec:efficiency}.
    \item $\mathrm{UR^3}$ also shares the inherent limitation associated with all the re-ranking approaches in that its maximum possible performance is dependent on the first-stage retrieval.
\end{itemize}
% % Future studies could explore the extension of $\mathrm{UR^3}$ to various domains and its potential to integrate with other AI systems, aiming for broader applicability and even greater accuracy in real-world information retrieval scenarios.

\section*{Acknowledgments}
This work is supported by the National Science and Technology Major Project (No. 2022ZD0116300) and the National Science Foundation of China (No. 62106249).

\bibliography{custom}
\clearpage

\appendix

\section{Bayes Decision Theory} \label{appendix:bayes}
A possible action of the re-ranking process involves reordering the document subset $\mathcal{C}$ to ensure that a document containing the correct answer is ranked as highly as possible.

In the general framework of Bayesian decision theory, each action $a$ is associated with a loss $L(a, \theta)$, which depends upon $\theta \equiv (\theta_{Q_e}, \{\theta_{D}\}^k, \{\theta'_{D}\}^k)$, including the query language model, document language models and estimated models based on a LLM. 
Based on the Figure~\ref{fig:framework}, a possible action is to return a single document $a = d$, and the loss function depends on $\theta_{Q_e}, \theta_{D}$ and $\theta'_{D}$, the \textbf{expected risk of action $a$} can be formulated as:
% The expected risk of action $a$ is given by
% $$R(a\mid \mathcal{U},\mathbf{q},\mathcal{S},\mathcal{C})=\int_{\Theta} L(a,\theta)p(\theta \mid \mathcal{U},\mathbf{q},\mathcal{S},\mathcal{C}) d\theta$$
% The Bayesian decision rule is then to present the document list $a^*$having the least expected risk:
% $$a^*=\underset{a}{\operatorname{arg\,min}} R(a|U,\mathbf{q},S,C)$$
% Now, if we assume that a possible action is to return a single document $a = d_i$, and that the loss function only depends on $\theta_{Q_e}, \theta_{D_i}$ and $\theta'_{D_i}$, the risk can be simplified to
\begin{align} 
R(&\mathbf{d}; \mathbf{q}) \stackrel{\text{def}}{=} R(a = \mathbf{d} \mid U, \mathbf{q}, \mathcal{S}, C, \theta') = \nonumber\\
& \int_{\theta_{Q_e}} \int_{\Theta'_D} \int_{\Theta_D} L(\theta_{Q_e}, \theta'_D, \theta_D) p(\theta_{Q_e} \mid \mathbf{q}, \mathcal{U}) \nonumber \\
&\quad\ \;\times  p(\theta'_D \mid \mathbf{d}, \theta') \, p(\theta_D \mid \mathbf{d}, \mathcal{S} ) ,d\theta_{Q_e} d\theta'_D \, d\theta_D \nonumber
\end{align}
% The distribution of $\theta'_D$ is directly given the conditional probability in Equation \ref{eq:defllm}. Here we reformulate it as $\hat\theta'_\mathbf{d}$ to keep consistency with the subsequent notation. % p=1
Instead of explicitly computing the parameter distributions, the value can be approximated at the posterior mode as follows: 
\begin{equation*} \label{eq:risk_analyse}
R(\mathbf{d}; \mathbf{q}) \propto L(\hat\theta_\mathbf{q}, \theta'_D,\hat\theta_\mathbf{d}) p(\hat\theta_\mathbf{q} \mid \mathbf{q}, \mathcal{U})(\hat\theta_\mathbf{d} \mid \mathbf{d}, \mathcal{S}) 
% &\qquad  p(\mathbf{d}, \theta') p(\hat\theta_\mathbf{d} \mid \mathbf{d}, \mathcal{S}) \nonumber
\end{equation*}
where the distribution of $\theta'_D$ is determined by the document $\mathbf{d}$ with LLM $\theta'$ as $p(\mathbf{d};\theta')$, thereby the posterior of $p(\theta'_D)$ is a point mass distribution. And
\begin{align*}
&\hat{\theta}_\mathbf{q} = \underset{\theta_{Q_e}}{\operatorname{arg\,max}} p(\theta_{Q_e} \mid \mathbf{q}, \mathcal{U}) \\
&\hat{\theta}_\mathbf{d} = \underset{\theta_D}{\operatorname{arg\,max}} p(\theta_D \mid \mathbf{d}, \mathcal{S}) 
\end{align*}

Based on the prior assumption in the QLM that the document prior $p(\mathbf{d})$ is uniform, we can infer that $p(\hat{\theta_d}| \mathbf{d}, \mathcal{S})$ is the same for all $\mathbf{d}$. 
% Consequently, the distribution $p(\mathbf{d}; \theta')$ is also uniform across all documents. 
For a specific query, the posterior distribution of the query model can also be dropped, because it is unrelated to the ranking of documents.

Hence, the formula of risk can be simplified as:
\begin{equation*}\label{eq:risk}
R(\mathbf{d}; \mathbf{q}) \propto L(\hat{\theta}_\mathbf{q}, \theta'_D,\hat{\theta}_\mathbf{d}) 
% &\propto \Delta(\hat{\theta}_\mathbf{q}, \hat{\theta'}_\mathbf{d}) + \frac{c_1}{c_2} \Delta(\hat{\theta}_\mathbf{d}, \hat{\theta'}_\mathbf{d}) \nonumber
\end{equation*}

To summarize, the document set $\mathcal{C}$ is represented through a series of $k$ sequential decisions. This process yields a list of documents ranked in ascending order according to the $R(\mathbf{d}; \mathbf{q})$. A smaller loss $L$ means a better ranking for the document.

\section{Distance-based Loss Function}\label{appendix:loss}
\begin{figure}[htbp]
    \centering
    \includegraphics[width=0.8\columnwidth]{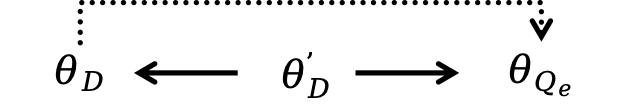}
    \caption{The estimations in $\mathrm{UR^3}$}
    \label{fig:append_upr3}
\end{figure}

\paragraph{KL framework of QLM.} The loss is calculated as: $$L(\theta_{Q_e},\theta_D) \propto \text{KL}[P(\theta_{Q_e})||P(\theta_D)]$$
The relevance value of a document with respect to a query is measured by the distance between two models. It is calculated by the KL divergence from the document model distribution $P(\theta_D)$ to the query model distribution $P(\theta_{Q_e})$. 
\begin{figure}[htbp]
    \centering
    \includegraphics[width=0.5\columnwidth]{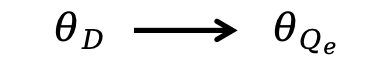}
    \caption{The estimation in QLM}
\end{figure}
% The matching problem is thus equivalent to measuring the $distance$ between the estimated query model and document model.
\paragraph{KL framwork of $\mathbf{UR^3}$.}
Based on the QLM framework, the calculation of $L(\theta_{Q_e}, \theta_D, \theta'_D)$ aims to measure the distance between the actual query and document model distributions through a LLM.
It can be interpreted as the proportional sum of the distance between the document model $\theta_D$ and the estimated model $\theta'_D$, and the distance from the estimated model $\theta'_D$ to the query model $\theta_{Q_e}$. 
We consider the two estimations are independent (left and right items in Figure~\ref{fig:append_upr3}), then we approximate the distance in QLM as the sum of the following items:
% , and the model $\theta'_D$ is estimated based on document model $\theta_D$. 
% Then the specific formula is as follows: $$L(\theta_{Q_e},\theta'_D,\theta_D) \propto \text{KL}[P(\theta_{Q_e},\theta_D)||P(\theta'_D)]$$
% Given $\theta_{Q_e}$ and $\theta_D$ are independent, their joint distribution, then
% $P(\theta_{Q_e},\theta_D)$ equals $P(\theta_{Q_e})P(\theta_D)$.
% % Independence between $\theta_{Q_e}$ and $\theta'_D$:: the model $\theta'_D$ acts in  such a way that its adjustment or estimation from $\theta_D$ does not influence its ability to estimate $\theta_Q$, and vice versa.
% % Then we can derive that
% \begin{align*}
%     & \quad \text{KL}[P(\theta_{Q_e},\theta_D)||P(\theta'_D)] \\
%     &= \sum_{\mathbf{q},\mathbf{d}} P(\theta_Q)P(\theta_D)  \log \frac{P(\theta_Q)P(\theta_D)}{P(\theta'_D)}
% \end{align*}

\begin{align*}
    &\quad L(\theta_{Q_e}, \theta_D, \theta'_D)\\
    & = c \cdot \text{KL}[P(\theta_{Q_e}||P(\theta_D)] \\
    % &= \sum_{\mathbf{q},\mathbf{d}} P(\theta_Q)P(\theta_D)  \log \frac{P(\theta_Q)P(\theta_D)}{P(\theta'_D)} \\
    % &= \sum_\mathbf{q} P(\theta_Q=\mathbf{q}) \log \frac{P(\theta_Q=\mathbf{q})}{P(\theta'_D)} \\
    % &\quad + \sum_\mathbf{d} P(\theta_D=\mathbf{d}) \log \frac{P(\theta_D=\mathbf{d})}{P(\theta'_D)} \\
    & \approx c_1\cdot \text{KL}[ P(\theta_Q) \parallel P(\theta'_D)] + c_1\cdot \text{KL}[P(\theta_D) \parallel P(\theta'_D))]
    % &= L(\theta_{Q_e},\theta'_D) +  L(\theta_D,\theta'_D) \\
    % &\approx c_1\Delta(\hat\theta_\mathbf{q}, \theta'_D) +c_2\Delta(\theta'_D, \hat\theta_\mathbf{d}) 
\end{align*}
where $c$, $c_1$ and $c_2$ are scale factors.

\section{Detailed Derivation} \label{appendix:derivation}
\subsection{Probability of Query Generation}\label{appendix:query}
Following the work of~\citet{conf/sigir/LaffertyZ01}, when the $\hat\theta_\mathbf{q}$ is considered as the empirical distribution of the query $\mathbf{q} = q_1 q_2 \ldots q_m$; that is,
\begin{equation*}
    p(w\mid \hat\theta_\mathbf{q})=-\frac{1}{m}\sum_{i=1}^{m} \delta (w,q_i)
\end{equation*}
where, $\delta$ is the indicator function, then we obtain
% Detailed derivation refers to~\citet{conf/sigir/LaffertyZ01}
\begin{align*} 
\Delta(\hat\theta_\mathbf{q}, \theta'_D) &\stackrel{\text{def}}{=} \text{KL}[\ p(\hat\theta_\mathbf{q}) \parallel p(\theta'_D)\ ] \\
& = \sum_w p(w\mid \hat\theta_\mathbf{q}) \log \frac{p(w\mid \hat\theta_\mathbf{q})}{p(w\mid \theta'_D)}\\
&\propto - \sum_w p(w\mid \hat\theta_\mathbf{q}) \log p(w\mid \theta'_D)+ c_\mathbf{q}\\
&\propto - (\sum_{w\in \mathbf{q}\cap D} \log p(w|\theta'_D) \\
&+ \sum_{w \in \mathbf{q}, w\notin D}\log p(w|\theta'_D))+ c_q \\
&\propto -\sum_{w\in \mathbf{q}} \log p(w|\theta'_D) + c_q \\
&\propto - \log p(\mathbf{q} \mid \theta'_D) + c_\mathbf{q} \\
&\propto -\frac{1}{m}\sum_{i=1}^{m} \log p(q_i \mid \mathbf{q}_{<i}, \mathbf{d};\theta') 
\end{align*}
where the constant $c_\mathbf{q}$ presents the entropy of the query model.
This is precisely the log-likelihood criterion that has been in used in all work on the language modeling of query generation approach.
% This is precisely the log-likelihood criterion that has been in used in the language modeling approaches of query generation~\cite{conf/emnlp/SachanLJAYPZ22,conf/emnlp/Zhuang0KZ23}.
\subsection{Evidence Lower Bound (ELBO)}\label{appendix:elbo}
Here we view $p(\hat\theta_\mathbf{d})$ and $p(\theta' \mid \mathbf{d})$ as two distributions across the space of $\theta$.
And we denote the distribution $p(\hat\theta_\mathbf{d})$ as $q(\theta)$ and $p(\theta' \mid \mathbf{d})$ as $p(\theta' \mid \mathbf{d})$, thus
\begin{align*}
& \quad \text{KL}[\ p(\hat\theta_\mathbf{d}) \parallel p(\theta' \mid \mathbf{d})]\\
&= \text{KL}(q(\theta) \parallel  p(\theta \mid \mathbf{d})) \\
&= - \int q(\theta) \log \frac{p(\theta \mid \mathbf{d})}{q(\theta)} \, d\theta \\
&= \int q(\theta) \log q(\theta) \, d\theta - \int q(\theta) \log p(\theta \mid \mathbf{d}) \, d\theta \\
&= \mathbb{E}_q[\log q(\theta)] - \mathbb{E}_q[\log p(\theta \mid \mathbf{d})] \\
&= \mathbb{E}_q[\log q(\theta)] - \mathbb{E}_q \left[ \log \frac{p(\mathbf{d}, \theta)}{p(\mathbf{d})} \right] \\
&= \mathbb{E}_q[\log q(\theta)] - \mathbb{E}_q[\log p(\mathbf{d}, \theta)] + \log p(\mathbf{d}) \\
&= -\text{ELBO}(\theta) + \log p(\mathbf{d}) 
\end{align*}
Then
\begin{align*}
&\quad\ \text{ELBO}(\theta) \\
&= \mathbb{E}_q[\log p(\mathbf{d}, \theta)] - \mathbb{E}_q[\log q(\theta)] \\
&= \mathbb{E}_q[\log p(\mathbf{d} \mid \theta) p(\theta)] - \mathbb{E}_q[\log q(\theta)] \\
&= \mathbb{E}_q[\log p(\mathbf{d} \mid \theta)] + \mathbb{E}_q[\log p(\theta)] - \mathbb{E}_q[\log q(\theta)] \\
&= \mathbb{E}_q[\log p(\mathbf{d} \mid \theta)] + \mathbb{E}_q \left[ \frac{\log p(\theta)}{\log q(\theta)} \right] \\
&= \mathbb{E}_q[\log p(\mathbf{d} \mid \theta)] + \int q(\theta) \frac{\log p(\theta)}{\log q(\theta)} \, d\theta \\
&= \mathbb{E}_q[\log p(\mathbf{d} \mid \theta)] - \text{KL}[\ q(\theta) \| p(\theta)] \\
& \approx \log p(\mathbf{d} \mid \theta') - \text{KL}[\ p(\hat\theta_\mathbf{d}) \parallel p(\theta')] 
\end{align*}
where the expectation of the term $\mathbb{E}_q[\log p(\mathbf{d} \mid \theta)]$ employs the generation probability on LLM $\theta'$ as $\log p(\mathbf{d} \mid \theta')$ to minimize computational costs.
\section{Query Generation Instruction}\label{sec:prompt}
The query generation instruction~\citep{sachan-etal-2021-end} uses the log-probability of the query.

% \newtcolorbox{toolmakerbox}[1]{colback=blue!5!white,colframe=blue!75!black,fonttitle=\bfseries,title=#1}

\begin{tcolorbox}
Please write a question based on this passage.

Passage: \code{\{\{passage\}\}}

Question: \code{\{\{query\}\}}
\end{tcolorbox}
\paragraph{Document Generation.}
Specifically, when calculating the probability of document generation, we compute the negative log loss using the document portion prior to the output query under the current prompt. This approach synchronizes the computation of the query and the document within the same output, significantly reducing computational costs.

\section{Performance on BEIR Benchmark} \label{appendix:beir}
To evaluate the generalization of our method, we conducted experiments a popular subset of the BEIR benchmark dataset~\cite{conf/nips/Thakur0RSG21}. The evaluation metrics employed are Top-1 accuracy and nDCG@10, the official metric for the BEIR benchmark.
For a fair comparison, all the re-rankers consider the Top 100 documents retrieved by Contriever. The results are shown in Table~\ref{tab:beir-benchmark}.

In summary, the results demonstrate the effectiveness of $\mathrm{UR^3}$ as the average Top-1 accuracy improves by 4.39\% and the NDCG@10 scores improve by 2.37\%.
Due to the diversity in datasets, there is a considerable variation in performance gains across them. 
% The highest relative performance gains are obtained by $\mathrm{UR^3}$ on datasets containing information-seeking questions such as Trec-Covid, NQ, DBpedia, etc. On other datasets, the relative gains from re-ranking are moderate to little. 
$\mathrm{UR^3}$ achieves the highest relative performance improvements on datasets such as Trec-Covid, NQ, Touche-2020 and DBpedia. These datasets typically involve information-seeking questions, which benefit significantly from our advanced re-ranking method. 
% For datasets like HotpotQA and MS-Marco, the gains are not as pronounced as those seen in the previously mentioned datasets. These datasets contain a mixture of simple and complex queries, allowing for moderate improvements through re-ranking.

We also observe a decline in performance on the FIQA, ArguAna, and Scidocs datasets, each characterized by high average document lengths. This suggests that these datasets contain more complex and extensive information, which could be challenging for $\mathrm{UR^3}$ to process effectively, since the $\mathrm{UR^3}$ might struggle with effectively calculating the estimation bias among documents with such complexity, causing a drop in performance. Additionally, the Finance and Science domains might pose specific challenges that $\mathrm{UR^3}$ is not optimized for.

% We anticipate that by experimenting with different prompt instructions in UPR to better suit the end-task and cross-validating with the number of top-K documents to be re-ranked, results can be improved on these datasets. However, we leave these explorations as a part of future work.
\renewcommand{\arraystretch}{0.92} 
\begin{table}[t]
\addtolength{\tabcolsep}{-0.45pt}
\tiny
\centering
\begin{tabular}{@{}l |c c c| c c c@{}}
 \toprule
 \multirow{3}{*}{\textbf{Dataset}}& \multicolumn{3}{c|}{\textbf{Top-1}} & \multicolumn{3}{c}{\textbf{NDCG@10}} \\
 \cmidrule{2-7}
  % & \multicolumn{2}{c}{\textbf{BM25}} & \multicolumn{2}{c|}{\textbf{Contriever}} & \multicolumn{2}{c}{\textbf{BM25}} & \multicolumn{2}{c}{\textbf{Contriever}} \\
   & Original & UPR & $\mathrm{UR^3}$ & Original & UPR & $\mathrm{UR^3}$   \\
% \midrule

% \multicolumn{7}{c}{\textit{Question Answering}} \\
\midrule
NQ            & 22.16 & 32.38 & 37.67 & 23.19 & 35.58 & 38.64 \\
HotpotQA      & 53.37 & 84.35 & 86.02 & 60.60 & 85.74 & 87.43 \\
FIQA     & 21.14 & 40.43 & \textcolor{red}{39.66} & 29.16 & 48.50 & \textcolor{red}{48.16} \\

% CQADupStack   & 29.9 & 41.6 & 28.4 & 41.7 & 60.6 & 70.1 \\
% \midrule
% \multicolumn{7}{c}{\textit{Passage Retrieval}} \\
% \midrule
MS-Marco      & \phantom{0}8.70 & 11.92 & 12.16 & 20.68 & 27.26 & 27.41 \\
Trec-Covid    & 44.00 & 64.00 & 72.00 & 33.43 & 62.22 & 66.77\\
% \textbf{Nfcorpus }     & 40.6 & 42.4 & 42.7 & 39.6 & 39.2 & 39.0 \\
% \midrule
% \multicolumn{7}{c}{\textit{Argument Retrieval}} \\
% \midrule
Touche-2020   & 22.49 & 10.23 & 30.64 & 23.89 & 19.68 & 27.03 \\
ArguAna       & \phantom{0}0.00 & \phantom{0}9.72 & \phantom{0}\textcolor{red}{7.33} & \phantom{0}0.31 & 28.38 & \textcolor{red}{22.91} \\
% \textbf{Quora}         & 74.7 & 70.3 & \textcolor{red}{69.9} & 83.7 & 80.9 & \textcolor{red}{80.6} \\

% \midrule
% \multicolumn{7}{c}{\textit{Entity Retrieval}} \\
% \midrule
DBpedia       & 48.25 & 43.00 & 50.75 & 37.64 & 40.37 & 44.65 \\
% \midrule
% \multicolumn{7}{c}{\textit{Fact Checking}} \\
% \midrule
Fever         & 52.51 & 41.13 & 48.23 & 70.33 & 53.32 & 60.14 \\
Climate-Fever & 12.64 & \phantom{0}7.88 & 12.18 & 20.68 & 15.18 & 21.04 \\
Scifact       & 51.71 & 54.33 & 55.72 & 65.04 & 64.78 & 65.43 \\
% \midrule
% \multicolumn{7}{c}{\textit{Citation Prediction}} \\
% \midrule
Scidocs       & 18.67 & 21.32 & \textcolor{red}{21.04} & 23.81 & 32.04 & \textcolor{red}{31.80} \\
\midrule
\midrule
\textbf{Average}       & 29.63 & 35.06 & \textbf{39.45} & 34.06 & 42.75 & \textbf{45.12} \\
\bottomrule
\end{tabular}
\caption{Re-ranking results on the Top100 documents retrieved by Contriever on BEIR benchmark~\cite{conf/nips/Thakur0RSG21}. On average, the performances improve both on the Top-1 accuracy and NDCG@10 metrics.
The drop in some datasets is highlighted in red.
}
\label{tab:beir-benchmark}
\end{table}
\renewcommand{\arraystretch}{1.0} 
\section{Re-ranking on Mistral-7B and GPT-Neo-2.7B models}
\label{appendix:reranking}
As illustrated in Table~\ref{tab:mistral-rerank} and \ref{tab:neo-rerank}, the results demonstrate that $\mathrm{UR^3}$ enhances the overall rankings of the Top-100 documents both on the Mistral and GPT-Neo models. The improvements of our method are observed across all nDCG@K metrics, indicating that $\mathrm{UR^3}$ prioritizes relevant documents more effectively compared to the UPR method.

\section{Document Distribution on Top-1}
\begin{figure}[hb]
    \centering
    \includegraphics[width=0.85\columnwidth]{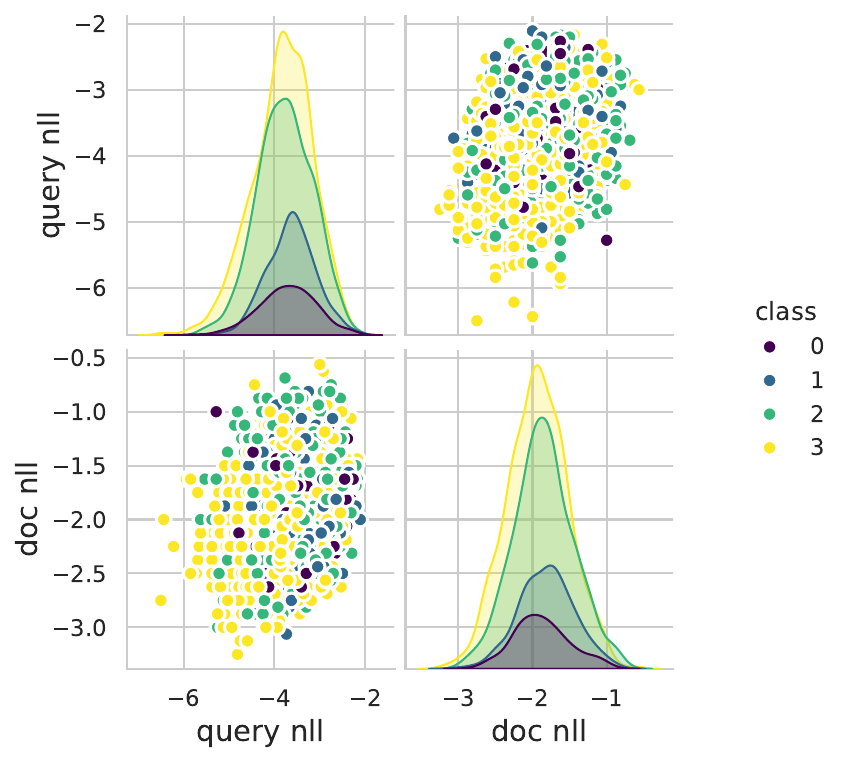}
    \caption{Complete results of Distribution on Top-1}
    \label{fig:scatter_all}
\end{figure}
The Figure~\ref{fig:scatter_all} presents the complete results of the scatter plot in Figure~\ref{fig:scatt}, which that statistically categorizes the relevant documents at the Top-1 rank, comparing UPR and UR3. The class 0 denotes a correct calibration where an irrelevant document ranked by UPR is adjusted to a relevant one at the Top-1 rank. The class 1 denotes a incorrect calibration from relevant document to irrelevant one. The class 2 denotes the correct results on both methods, while class 3 representes the incorrect results on both models.

\section{Discussion on Hyperparameters} \label{appendix:hyper}
Here we display the detailed results about hyperparameter $\alpha$ analysis on different datasets of LLaMA2-7B model with nDCG@K metrics. 
% As illustrated in Figure~\ref{fig:hyper}, we find $\alpha=0.25$ provide robust improvements over other values mostly. Although the best value of NDCG@1 on the webq dataset is larger than 0.25, the comprehensive indicators are still the best choice 0.25.
% Therefore, we adopt this hyperparameter

\begin{figure}[t]
\centering
    \begin{subfigure}[b]{0.24\textwidth}
        \includegraphics[width=\textwidth]{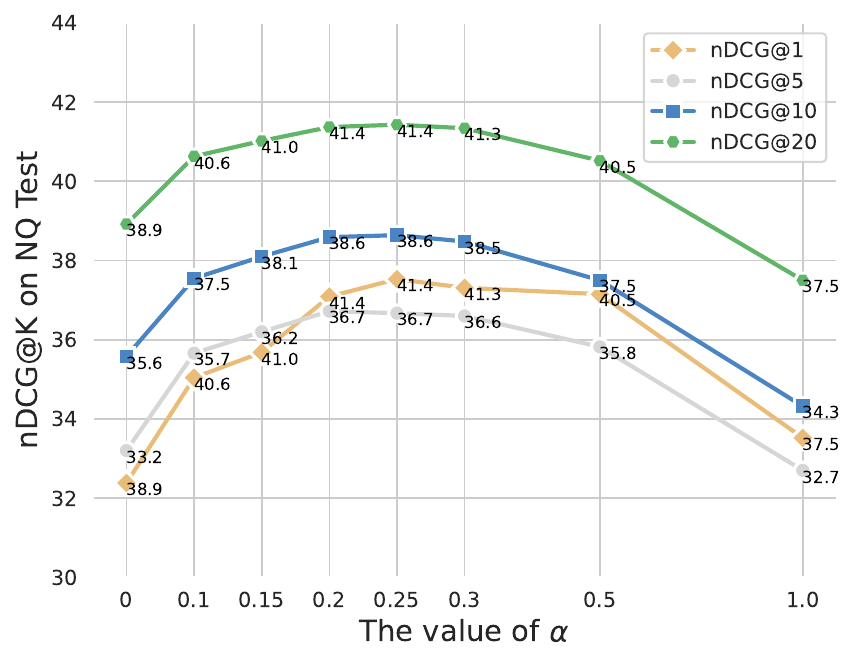}
  \caption{NQ}
  \label{fig:hyper_nq}
\end{subfigure}%
\hfill % Adds horizontal space between the figures
    \begin{subfigure}[b]{0.24\textwidth}
        \includegraphics[width=\textwidth]{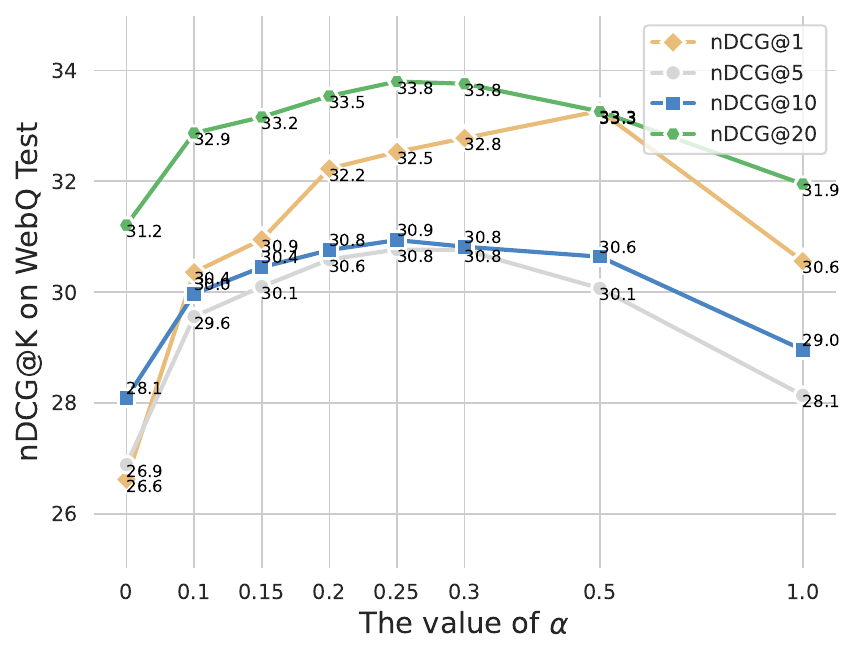}
  \caption{WebQ}
  \label{fig:hyper_webq}
\end{subfigure}
\caption{Comparative effects of the hyperparameter on nDCG@K across different datasets. Evaluation is done on the Contriever retrieved documents.}
\end{figure}

As depicted in Figure~\ref{fig:hyper_nq} and \ref{fig:hyper_webq}, our analysis reveals that a hyperparameter setting of $\alpha=0.25$ consistently yields robust enhancements compared to other evaluated values. While the highest observed NDCG@1 on the WebQ dataset exceeds 0.25, the overall performance metrics substantiate that 0.25 remains the optimal choice. Consequently, this hyperparameter configuration is adopted across all experimental evaluations.

% \section{QA with different number of documents on LLaMA2-7B model}
% As illustrated in Table, we conduct 

\section{Performance on Paraphrased Query}
To evaluate the retrieval outcomes on paraphrased queries, we utilized the GPT-3.5 Turbo model to paraphrase the queries from the WebQ dataset. The paraphrasing process was guided by the prompt, \textit{"Please paraphrase the following question: {question}"}. Below are examples of the paraphrased questions we generated:
\begin{itemize}
    \item Original Question: "What happened after Mr. Sugihara died?"-> Paraphrased: "What occurred following Mr. Sugihara's passing?"
    \item Original Question: "Where was George Washington Carver from?"-> Paraphrased: "What was George Washington Carver's place of origin?"
    \item Original Question: "Who was Richard Nixon married to?"-> Paraphrased: "Who was the spouse of Richard Nixon?"
\end{itemize}
We then assessed the re-ranking results across various retrievers, including Contriever, BM25, MSS, and DPR, utilizing the paraphrased queries.

\begin{table}[h]
\centering
\tiny
\begin{tabular}{lcccccc}
\hline
\textbf{Llama2-7b} & \textbf{Contriever} & \multicolumn{2}{c}{\textbf{UPR}} & \multicolumn{2}{c}{\textbf{$UR^3$}} \\
\textbf{Metric}& & \textbf{w/o para.} & \textbf{w/ para.} & \textbf{w/o para.} & \textbf{w/ para.} \\ \hline
top1           & 19.98  & 26.62 & 28.44 & 32.53 & 33.71 \\
top5           & 43.45  & 54.92 & 56.35 & 58.71 & 59.30  \\
top20          & 65.70   & 72.69 & 71.75(↓) & 73.43 & 73.61 \\
\hline
ndcg@1         & 19.98  & 26.62 & 28.44 & 32.53 & 33.71 \\
ndcg@5         & 18.64  & 26.78 & 28.27 & 30.82 & 31.51 \\
ndcg@20        & 22.22  & 31.18 & 32.24 & 33.79 & 34.61 \\
\hline
MAP@100        & 18.79  & 25.92 & 27.12 & 27.82 & 28.67 \\ \hline
\end{tabular}
\caption{Performance comparison on Contriever}
\end{table}

\begin{table}[h]
\centering
\tiny
\begin{tabular}{lcccccc}
\hline
\textbf{Llama2-7b} & \textbf{BM25} & \multicolumn{2}{c}{\textbf{UPR}} & \multicolumn{2}{c}{\textbf{$UR^3$}} \\
\textbf{Metric}& & \textbf{w/o para.} & \textbf{w/ para.} & \textbf{w/o para.} & \textbf{w/ para.} \\ \hline
top1           & 18.90   & 27.56 & 28.25 & 33.91 & 34.10  \\
top5           & 41.83  & 54.13 & 54.28 & 55.17 & 56.35 \\
top20          & 62.40   & 68.5  & 69.05 & 69.54 & 69.98 \\
\hline
ndcg@1         & 18.90   & 27.56 & 28.25 & 33.91 & 34.10  \\
ndcg@5         & 19.36  & 27.39 & 28.26 & 30.72 & 31.34 \\
ndcg@20        & 22.12  & 31.44 & 32.18 & 33.62 & 34.07 \\
\hline
MAP@100        & 19.15  & 26.63 & 27.18 & 28.09 & 28.45 \\ \hline
\end{tabular}
\caption{Performance comparison on BM25}
\end{table}
\begin{table}[h]
\centering
\tiny
\begin{tabular}{lcccccc}
\hline
\textbf{Llama2-7b} & \textbf{MSS} & \multicolumn{2}{c}{\textbf{UPR}} & \multicolumn{2}{c}{\textbf{$UR^3$}} \\
\textbf{Metric}& & \textbf{w/o para.} & \textbf{w/ para.} & \textbf{w/o para.} & \textbf{w/ para.} \\
\hline
top1           & 11.66 & 26.38 & 26.03(↓) & 29.38 & 29.72 \\
top5           & 29.04 & 48.67 & 50.34 & 49.85 & 51.48 \\
top20          & 49.21 & 63.19 & 62.80(↓)  & 62.40  & 62.80  \\
\hline
ndcg@1         & 11.66 & 26.38 & 26.03 & 29.38 & 29.72 \\
ndcg@5         & 11.57 & 26.67 & 27.47 & 28.21 & 29.04 \\
ndcg@20        & 14.84 & 32.46 & 32.77 & 33.20  & 33.61 \\
\hline
MAP@100        & 12.03 & 26.20  & 26.67 & 26.84 & 27.14 \\ \hline

\end{tabular}
\caption{Performance comparison on MSS}
\end{table}
\begin{table}[h]
\centering
\tiny
\begin{tabular}{lcccccc}
\hline
\textbf{Llama2-7b} & \textbf{DPR} & \multicolumn{2}{c}{\textbf{UPR}} & \multicolumn{2}{c}{\textbf{$UR^3$}} \\
\textbf{Metric}& & \textbf{w/o para.} & \textbf{w/ para.} & \textbf{w/o para.} & \textbf{w/ para.} \\ \hline
top1           & 44.83 & 39.32 & 37.89(↓) & 42.18 & 42.86 \\
top5           & 65.01 & 66.83 & 64.76(↓) & 66.88 & 66.91 \\
top20          & 74.61 & 76.67 & 76.53(↓) & 76.96 & 76.93 \\
\hline
ndcg@1         & 44.83 & 39.32 & 37.89(↓) & 42.18 & 42.86 \\
ndcg@5         & 39.76 & 38.66 & 37.95(↓) & 40.34 & 40.97 \\
ndcg@20        & 38.95 & 41.81 & 41.93 & 42.65 & 43.10  \\
\hline
MAP@100        & 33.32 & 36.46 & 36.63 & 36.82 & 37.44 \\ \hline
\end{tabular}
\caption{Performance comparison on DPR}
\end{table}

We observed that the queries paraphrased using ChatGPT generally resulted in marginal improvements in the reranking of retrieval results both on UPR and $\mathrm{UR^3}$ for Contriever, BM25, and MSS. Our analysis suggests that this outcome may stem from the lower accuracy of these retrieval results to begin with. Furthermore, the distribution of queries generated by ChatGPT aligns more closely with the model data distribution than the empirical distribution of the original queries. This alignment potentially offers beneficial support for reranking initially poor retrieval outcomes.

However, for supervised retrieval systems like DPR, which already achieve high accuracy, the paraphrased queries deviate from the empirical data distribution, leading to greater negative impacts. Consequently, the performance of the UPR method is noticeably affected.

Nonetheless, our approach involves a bias-corrected estimation, which, by leveraging the document probability values, mitigates the performance decline observed in DPR results and even achieves slight improvements.

\section{More Experiments on QA Task}
We conduct inference on the re-ranking results of Mistral-7B in Table~\ref{tab:odqa-mistral}. The $\mathrm{UR^3}$ method substantially enhances the performance of QA tasks, achieving superior EM and F1 scores compared to the UPR method on Mistral model. While DPR method has better performances on NQ and WebQ datasets for LLaMA2-13B and Gemma-7B, this trend is consistent with the analysis in Section~\ref{exp:analysis}.

\begin{table*}[hb]
    \small
    \centering
    \resizebox{\textwidth}{!}{%
    \begin{threeparttable}
    \begin{tabular}{llccc|ccc|ccc|ccc}
         \cmidrule[1pt]{1-14}
         % & & \multicolumn{9}{c}{\textit{Unsupervised Retrievers}} & \multicolumn{3}{c}{\textit{Supervised Retrievers}} \\
         & & \multicolumn{3}{c}{\textbf{Contriever}} & \multicolumn{3}{c}{\textbf{BM25}} & \multicolumn{3}{c}{\textbf{MSS}} & \multicolumn{3}{c}{\textbf{DPR}}  \\
         \cmidrule(lr){3-5} \cmidrule(lr){6-8} \cmidrule(lr){9-11} \cmidrule(lr){12-14} %\cmidrule(lr){15-17}
         \textbf{Datasets} &\textbf{Metric} & Orig. & UPR & $\mathrm{UR^3}$ &  Orig. & UPR &  $\mathrm{UR^3}$ &  Orig. & UPR &   $\mathrm{UR^3}$ &  Orig. & UPR & $\mathrm{UR^3}$  \\
         \midrule
         \multirow{7}{*}{NQ}& Top-1 & \cellcolor{gray!10}22.16 & 32.63 & \textbf{38.61} & \cellcolor{gray!10}22.11 & 32.55 & \textbf{37.89} & \cellcolor{gray!10}19.28 & 32.60 & \textbf{37.04} & \cellcolor{gray!10}46.34  & 37.65 & \textbf{43.30}  \\
        % & Top-3 & \cellcolor{gray!10}36.59 & 52.11 & 54.74 & \cellcolor{gray!10}36.59 & 52.11 & 54.74 & \cellcolor{gray!10}0.91 & 0.88 & 0.84 & \cellcolor{gray!10}0.78 & 0.71 & 0.82 \\
        & Top-5 & \cellcolor{gray!10}47.26 & 61.91 & \textbf{64.82} & \cellcolor{gray!10}43.77 & 60.36 & \textbf{63.24} & \cellcolor{gray!10}41.25 & 59.72 & \textbf{61.75} & \cellcolor{gray!10}68.28  & 68.73 & \textbf{72.27}  \\
        % & Top-10 & \cellcolor{gray!10}54.46 & 67.26 & 68.50 & \cellcolor{gray!10}54.46 & 67.26 & 68.50 & \cellcolor{gray!10}0.91 & 0.88 & 0.84 & \cellcolor{gray!10}0.78 & 0.71 & 0.82 \\
        & Top-20 & \cellcolor{gray!10}67.87 & 75.57 & \textbf{76.76} & \cellcolor{gray!10}62.94 & 72.77 & \textbf{73.52} & \cellcolor{gray!10}59.97 & 71.25 & \textbf{71.61} & \cellcolor{gray!10}80.06  & 81.99 & \textbf{82.77}  \\
        % & Top-50 & \cellcolor{gray!10}0.86 & 0.86 & 0.86 & \cellcolor{gray!10}72.74 & 76.87 & 76.87 & \cellcolor{gray!10}0.91 & 0.88 & 0.84 & \cellcolor{gray!10}0.78 & 0.71 & 0.82 &  \\
        \cmidrule(lr){2-14}
        & nDCG@1 & \cellcolor{gray!10}22.16 & 32.63 & \textbf{38.61} & \cellcolor{gray!10}22.11 & 32.55 & \textbf{37.89} & \cellcolor{gray!10}19.28 & 32.60 & \textbf{37.04} & \cellcolor{gray!10}46.34  & 37.65 & \textbf{43.30}  \\
        % & nDCG@3 & \cellcolor{gray!10}21.56 & 32.90 & 36.83 & \cellcolor{gray!10}21.56 & 32.90 & 36.83 & \cellcolor{gray!10}0.91 & 0.88 & 0.84 & \cellcolor{gray!10}0.78 & 0.71 & 0.82  \\
        & nDCG@5 & \cellcolor{gray!10}21.70 & 33.67 & \textbf{38.21}& \cellcolor{gray!10}21.63 & 34.02 & \textbf{38.03} & \cellcolor{gray!10}18.97 & 34.53 & \textbf{37.90} & \cellcolor{gray!10}40.62  & 38.37 & \textbf{43.21}  \\
        % & nDCG@10 & \cellcolor{gray!10}23.10 & 35.76 & 38.27 & \cellcolor{gray!10}23.10 & 35.76 & 38.27 & \cellcolor{gray!10}0.91 & 0.88 & 0.84 & \cellcolor{gray!10}0.78 & 0.71 & 0.82 \\
        & nDCG@20 & \cellcolor{gray!10}26.15 & 39.02 & \textbf{42.66} & \cellcolor{gray!10}25.75 & 39.57 & \textbf{42.34} & \cellcolor{gray!10}22.88 & 39.43 & \textbf{41.81} & \cellcolor{gray!10}42.42  & 44.32 & \textbf{47.85}  \\
        % & nDCG@50 & \cellcolor{gray!10}25.75 & 39.70 & 41.15 & \cellcolor{gray!10}32.48 & 44.81 & 46.69 & \cellcolor{gray!10}0.91 & 0.88 & 0.84 & \cellcolor{gray!10}0.78 & 0.71 & 0.82 \\
        % & nDCG@100 & \cellcolor{gray!10}25.75 & 39.70 & 41.15 & \cellcolor{gray!10}40.12 & 49.27 & 50.94 & \cellcolor{gray!10}0.91 & 0.88 & 0.84 & \cellcolor{gray!10}0.78 & 0.71 & 0.82  \\
        \cmidrule(lr){2-14}
        & MAP@100 & \cellcolor{gray!10}20.71 & 31.65 & \textbf{35.06} & \cellcolor{gray!10}20.78 & 32.36 & \textbf{35.02} & \cellcolor{gray!10}18.11 & 32.41 & \textbf{34.77} & \cellcolor{gray!10}34.89  & 36.34 & \textbf{39.54}  \\
         % \bottomrule
        \toprule[0.5pt]
         \multirow{7}{*}{WebQ}& Top-1 & \cellcolor{gray!10}19.98 & 28.44 & \textbf{33.37} & \cellcolor{gray!10}18.90 & 29.08 & \textbf{33.61} & \cellcolor{gray!10}11.66 & 27.31 & \textbf{30.12} & \cellcolor{gray!10}\textbf{44.83}  & 39.03 & 43.06  \\
        % & Top-3 & \cellcolor{gray!10}36.59 & 52.11 & 54.74 & \cellcolor{gray!10}36.59 & 52.11 & 54.74 & \cellcolor{gray!10}0.91 & 0.88 & 0.84 & \cellcolor{gray!10}0.78 & 0.71 & 0.82 \\
        & Top-5 & \cellcolor{gray!10}43.45 & 56.25 & \textbf{60.86} & \cellcolor{gray!10}41.83 & 54.13 & \textbf{55.95} & \cellcolor{gray!10}29.04 & 49.56 & \textbf{51.13} & \cellcolor{gray!10}65.01  & 66.58 & \textbf{67.96}  \\
        % & Top-10 & \cellcolor{gray!10}54.46 & 67.26 & 68.50 & \cellcolor{gray!10}54.46 & 67.26 & 68.50 & \cellcolor{gray!10}0.91 & 0.88 & 0.84 & \cellcolor{gray!10}0.78 & 0.71 & 0.82 \\
        & Top-20 & \cellcolor{gray!10}65.70 & 72.39 & \textbf{73.67} & \cellcolor{gray!10}62.40 & 68.80 & \textbf{69.54} & \cellcolor{gray!10}49.21 &\textbf{ 62.89} & 62.50 & \cellcolor{gray!10}74.61  & 76.57 & \textbf{77.17}  \\
        % & Top-50 & \cellcolor{gray!10}0.86 & 0.86 & 0.86 & \cellcolor{gray!10}72.74 & 76.87 & 76.87 & \cellcolor{gray!10}0.91 & 0.88 & 0.84 & \cellcolor{gray!10}0.78 & 0.71 & 0.82 &  \\
        \cmidrule(lr){2-14}
        & nDCG@1 & \cellcolor{gray!10}19.98 & 28.44 & \textbf{33.37} & \cellcolor{gray!10}18.90 & 29.08 & \textbf{33.61} & \cellcolor{gray!10}11.66 & 27.31 & \textbf{30.12} & \cellcolor{gray!10}\textbf{44.83}  & 39.03 & 43.06  \\
        % & nDCG@3 & \cellcolor{gray!10}21.56 & 32.90 & 36.83 & \cellcolor{gray!10}21.56 & 32.90 & 36.83 & \cellcolor{gray!10}0.91 & 0.88 & 0.84 & \cellcolor{gray!10}0.78 & 0.71 & 0.82  \\
        & nDCG@5 & \cellcolor{gray!10}18.64 & 27.88 & \textbf{31.47} & \cellcolor{gray!10}19.36 & 28.26 & \textbf{31.47} & \cellcolor{gray!10}11.57 & 27.36 & \textbf{29.53} & \cellcolor{gray!10}39.76  & 39.14 & \textbf{41.07}  \\
        % & nDCG@10 & \cellcolor{gray!10}23.10 & 35.76 & 38.27 & \cellcolor{gray!10}23.10 & 35.76 & 38.27 & \cellcolor{gray!10}0.91 & 0.88 & 0.84 & \cellcolor{gray!10}0.78 & 0.71 & 0.82 \\
        & nDCG@20 & \cellcolor{gray!10}22.22 & 31.70 & \textbf{34.59} & \cellcolor{gray!10}22.12 & 32.01 & \textbf{34.15} & \cellcolor{gray!10}14.84 & 32.97 & \textbf{34.16} & \cellcolor{gray!10}38.95  & 42.16 & \textbf{43.39}  \\
        % & nDCG@50 & \cellcolor{gray!10}25.75 & 39.70 & 41.15 & \cellcolor{gray!10}32.48 & 44.81 & 46.69 & \cellcolor{gray!10}0.91 & 0.88 & 0.84 & \cellcolor{gray!10}0.78 & 0.71 & 0.82 \\
        % & nDCG@100 & \cellcolor{gray!10}25.75 & 39.70 & 41.15 & \cellcolor{gray!10}40.12 & 49.27 & 50.94 & \cellcolor{gray!10}0.91 & 0.88 & 0.84 & \cellcolor{gray!10}0.78 & 0.71 & 0.82  \\
        \cmidrule(lr){2-14}
        & MAP@100 & \cellcolor{gray!10}18.79 & 26.31 & \textbf{28.49} & \cellcolor{gray!10}19.15 & 26.98 & \textbf{28.59} & \cellcolor{gray!10}12.03 & 26.72 & \textbf{28.08} & \cellcolor{gray!10}33.32  & 36.63 & \textbf{37.53}  \\
        \toprule[0.5pt]
        \multirow{7}{*}{TriviaQA}& Top-1 & \cellcolor{gray!10}34.16 & 52.63 & \textbf{56.07} & \cellcolor{gray!10}46.30 & 55.48 & \textbf{58.24} & \cellcolor{gray!10}30.76 & 52.87 & \textbf{55.00} & \cellcolor{gray!10}57.47  & 62.48 & \textbf{63.99}  \\
        % & Top-3 & \cellcolor{gray!10}36.59 & 52.11 & 54.74 & \cellcolor{gray!10}36.59 & 52.11 & 54.74 & \cellcolor{gray!10}0.91 & 0.88 & 0.84 & \cellcolor{gray!10}0.78 & 0.71 & 0.82 \\
        & Top-5 & \cellcolor{gray!10}59.49 & 73.99 & \textbf{74.75} & \cellcolor{gray!10}66.28 & 75.42 & \textbf{75.89} & \cellcolor{gray!10}52.65 & 70.64 & \textbf{71.05} & \cellcolor{gray!10}72.40  & \textbf{79.08} & 79.04  \\
        % & Top-10 & \cellcolor{gray!10}54.46 & 67.26 & 68.50 & \cellcolor{gray!10}54.46 & 67.26 & 68.50 & \cellcolor{gray!10}0.91 & 0.88 & 0.84 & \cellcolor{gray!10}0.78 & 0.71 & 0.82 \\
        & Top-20 & \cellcolor{gray!10}73.91 & 79.83 & \textbf{80.25} & \cellcolor{gray!10}76.41 & 80.77 & \textbf{80.87} & \cellcolor{gray!10}67.18 & 76.31 & \textbf{76.34} & \cellcolor{gray!10}79.77  & \textbf{83.13} & 83.09  \\
        % & Top-50 & \cellcolor{gray!10}0.86 & 0.86 & 0.86 & \cellcolor{gray!10}72.74 & 76.87 & 76.87 & \cellcolor{gray!10}0.91 & 0.88 & 0.84 & \cellcolor{gray!10}0.78 & 0.71 & 0.82 &  \\
        \cmidrule(lr){2-14}
        & nDCG@1 & \cellcolor{gray!10}34.16 & 52.63 & \textbf{56.07} & \cellcolor{gray!10}46.30 & 55.48 & \textbf{58.24} & \cellcolor{gray!10}30.76 & 52.87 & \textbf{55.00} & \cellcolor{gray!10}57.47  & 62.48 & \textbf{63.99}  \\
        % & nDCG@3 & \cellcolor{gray!10}21.56 & 32.90 & 36.83 & \cellcolor{gray!10}21.56 & 32.90 & 36.83 & \cellcolor{gray!10}0.91 & 0.88 & 0.84 & \cellcolor{gray!10}0.78 & 0.71 & 0.82  \\
        & nDCG@5 & \cellcolor{gray!10}30.46 & 49.63 & \textbf{51.78} & \cellcolor{gray!10}41.60 & 53.35 & \textbf{55.17} & \cellcolor{gray!10}27.78 & 50.64 & \textbf{52.04} & \cellcolor{gray!10}49.69  & 59.60 & \textbf{60.31} \\
        % & nDCG@10 & \cellcolor{gray!10}23.10 & 35.76 & 38.27 & \cellcolor{gray!10}23.10 & 35.76 & 38.27 & \cellcolor{gray!10}0.91 & 0.88 & 0.84 & \cellcolor{gray!10}0.78 & 0.71 & 0.82 \\
        & nDCG@20 & \cellcolor{gray!10}31.78 & 51.10 & \textbf{52.61} & \cellcolor{gray!10}40.68 & 54.72 & \textbf{55.88} & \cellcolor{gray!10}29.25 & 53.22 & \textbf{54.10} & \cellcolor{gray!10}46.33  & 60.05 & \textbf{60.27}  \\
        % & nDCG@50 & \cellcolor{gray!10}25.75 & 39.70 & 41.15 & \cellcolor{gray!10}32.48 & 44.81 & 46.69 & \cellcolor{gray!10}0.91 & 0.88 & 0.84 & \cellcolor{gray!10}0.78 & 0.71 & 0.82 \\
        % & nDCG@100 & \cellcolor{gray!10}25.75 & 39.70 & 41.15 & \cellcolor{gray!10}40.12 & 49.27 & 50.94 & \cellcolor{gray!10}0.91 & 0.88 & 0.84 & \cellcolor{gray!10}0.78 & 0.71 & 0.82  \\
        \cmidrule(lr){2-14}
        & MAP@100 & \cellcolor{gray!10}26.61 & 44.86 & \textbf{46.06} & \cellcolor{gray!10}34.85 & 49.36 & \textbf{50.36} & \cellcolor{gray!10}24.02 & 47.12 & \textbf{47.95} & \cellcolor{gray!10}39.40  & 54.25 & \textbf{54.46}  \\
        \bottomrule[1pt]
    \end{tabular}
     \end{threeparttable}
    }
    \vspace{-3mm}
    \caption{Re-ranking results on the test set of datasets of the Top-100 retrieved documents with the Mistral-7B model. The best results are highlighted in bold.} \label{tab:mistral-rerank}
    \vspace{-3mm}
\end{table*}

% \section{The Performances of GPT-Neo}

\begin{table*}[hb]
    \small
    \centering
    \resizebox{\textwidth}{!}{%
    \begin{threeparttable}
    \begin{tabular}{llccc|ccc|ccc|ccc}
         \cmidrule[1pt]{1-14}
         % & & \multicolumn{9}{c}{\textit{Unsupervised Retrievers}} & \multicolumn{3}{c}{\textit{Supervised Retrievers}} \\
         & & \multicolumn{3}{c}{\textbf{Contriever}} & \multicolumn{3}{c}{\textbf{BM25}} & \multicolumn{3}{c}{\textbf{MSS}} & \multicolumn{3}{c}{\textbf{DPR}}  \\
         \cmidrule(lr){3-5} \cmidrule(lr){6-8} \cmidrule(lr){9-11} \cmidrule(lr){12-14} %\cmidrule(lr){15-17}
         \textbf{Datasets} &\textbf{Metric} & Orig. & UPR & $\mathrm{UR^3}$ &  Orig. & UPR &  $\mathrm{UR^3}$ &  Orig. & UPR &   $\mathrm{UR^3}$ &  Orig. & UPR & $\mathrm{UR^3}$  \\
         \midrule
         \multirow{7}{*}{NQ}& Top-1 & \cellcolor{gray!10}22.16 & 29.86 & \textbf{34.02} & \cellcolor{gray!10}22.11 & 29.83 & \textbf{33.77} & \cellcolor{gray!10}19.28 & 30.78 & \textbf{33.63} & \cellcolor{gray!10}\textbf{46.34}  & 36.48 & 40.64  \\
        % & Top-3 & \cellcolor{gray!10}36.59 & 52.11 & 54.74 & \cellcolor{gray!10}36.59 & 52.11 & 54.74 & \cellcolor{gray!10}0.91 & 0.88 & 0.84 & \cellcolor{gray!10}0.78 & 0.71 & 0.82 \\
        & Top-5 & \cellcolor{gray!10}47.29 & 57.45 & \textbf{59.72} & \cellcolor{gray!10}43.77 & 56.34 & \textbf{58.31} & \cellcolor{gray!10}41.25 & 56.34 & \textbf{57.92} & \cellcolor{gray!10}68.28  & 66.90 & \textbf{68.70}  \\
        % & Top-10 & \cellcolor{gray!10}54.46 & 67.26 & 68.50 & \cellcolor{gray!10}54.46 & 67.26 & 68.50 & \cellcolor{gray!10}0.91 & 0.88 & 0.84 & \cellcolor{gray!10}0.78 & 0.71 & 0.82 \\
        & Top-20 & \cellcolor{gray!10}67.87 & 74.16 & \textbf{74.65} & \cellcolor{gray!10}62.94 & 71.63 & \textbf{71.69} & \cellcolor{gray!10}59.97 & 69.86 & \textbf{69.86} & \cellcolor{gray!10}80.06  & 81.16 & \textbf{81.99}  \\
        % & Top-50 & \cellcolor{gray!10}0.86 & 0.86 & 0.86 & \cellcolor{gray!10}72.74 & 76.87 & 76.87 & \cellcolor{gray!10}0.91 & 0.88 & 0.84 & \cellcolor{gray!10}0.78 & 0.71 & 0.82 &  \\
        \cmidrule(lr){2-14}
        & nDCG@1 & \cellcolor{gray!10}22.16 & 29.86 & \textbf{34.02} & \cellcolor{gray!10}22.11 & 29.83 & \textbf{33.77} & \cellcolor{gray!10}19.28 & 30.78 & \textbf{33.63} & \cellcolor{gray!10}\textbf{46.34}  & 36.48 & 40.64  \\
        % & nDCG@3 & \cellcolor{gray!10}21.56 & 32.90 & 36.83 & \cellcolor{gray!10}21.56 & 32.90 & 36.83 & \cellcolor{gray!10}0.91 & 0.88 & 0.84 & \cellcolor{gray!10}0.78 & 0.71 & 0.82  \\
        & nDCG@5 & \cellcolor{gray!10}21.70 & 30.93 & \textbf{33.58} & \cellcolor{gray!10}21.63 & 31.32 & \textbf{33.55} & \cellcolor{gray!10}18.97 & 32.06 & \textbf{34.07} & \cellcolor{gray!10}\textbf{40.62}  & 37.34 & 39.87  \\
        % & nDCG@10 & \cellcolor{gray!10}23.10 & 35.76 & 38.27 & \cellcolor{gray!10}23.10 & 35.76 & 38.27 & \cellcolor{gray!10}0.91 & 0.88 & 0.84 & \cellcolor{gray!10}0.78 & 0.71 & 0.82 \\
        & nDCG@20 & \cellcolor{gray!10}26.15 & 35.97 & \textbf{37.99} & \cellcolor{gray!10}25.75 & 36.65 & \textbf{38.27} & \cellcolor{gray!10}22.88 & 36.99 & \textbf{38.23} & \cellcolor{gray!10}42.42  & 42.43 & \textbf{44.39}  \\
        % & nDCG@50 & \cellcolor{gray!10}25.75 & 39.70 & 41.15 & \cellcolor{gray!10}32.48 & 44.81 & 46.69 & \cellcolor{gray!10}0.91 & 0.88 & 0.84 & \cellcolor{gray!10}0.78 & 0.71 & 0.82 \\
        % & nDCG@100 & \cellcolor{gray!10}25.75 & 39.70 & 41.15 & \cellcolor{gray!10}40.12 & 49.27 & 50.94 & \cellcolor{gray!10}0.91 & 0.88 & 0.84 & \cellcolor{gray!10}0.78 & 0.71 & 0.82  \\
        \cmidrule(lr){2-14}
        & MAP@100 & \cellcolor{gray!10}20.71 & 29.17 & \textbf{30.92} & \cellcolor{gray!10}20.78 & 30.11 & \textbf{31.48} & \cellcolor{gray!10}18.11 & 30.39 & \textbf{31.49} & \cellcolor{gray!10}34.89  & 34.86 & \textbf{36.43}  \\
         % \bottomrule
        \toprule[0.5pt]
         \multirow{7}{*}{WebQ}& Top-1 & \cellcolor{gray!10}19.98 & 26.13 & \textbf{28.40} & \cellcolor{gray!10}18.90 & 27.21 & \textbf{29.82} & \cellcolor{gray!10}11.66 & 25.39 & \textbf{28.00} & \cellcolor{gray!10}\textbf{44.83}  & 36.86 & 39.62  \\
        % & Top-3 & \cellcolor{gray!10}36.59 & 52.11 & 54.74 & \cellcolor{gray!10}36.59 & 52.11 & 54.74 & \cellcolor{gray!10}0.91 & 0.88 & 0.84 & \cellcolor{gray!10}0.78 & 0.71 & 0.82 \\
        & Top-5 & \cellcolor{gray!10}43.45 & 54.53 & \textbf{57.33} & \cellcolor{gray!10}41.83 & 52.51 & \textbf{52.95} & \cellcolor{gray!10}29.04 & 49.26 & \textbf{50.49} & \cellcolor{gray!10}65.01  & 63.73 & \textbf{65.70}  \\
        % & Top-10 & \cellcolor{gray!10}54.46 & 67.26 & 68.50 & \cellcolor{gray!10}54.46 & 67.26 & 68.50 & \cellcolor{gray!10}0.91 & 0.88 & 0.84 & \cellcolor{gray!10}0.78 & 0.71 & 0.82 \\
        & Top-20 & \cellcolor{gray!10}65.70 & 71.36 & \textbf{72.74} & \cellcolor{gray!10}62.40 & 68.01 & \textbf{68.26} & \cellcolor{gray!10}49.21 & 62.16 & \textbf{62.20} & \cellcolor{gray!10}74.61  & 75.74 & \textbf{76.23}  \\
        % & Top-50 & \cellcolor{gray!10}0.86 & 0.86 & 0.86 & \cellcolor{gray!10}72.74 & 76.87 & 76.87 & \cellcolor{gray!10}0.91 & 0.88 & 0.84 & \cellcolor{gray!10}0.78 & 0.71 & 0.82 &  \\
        \cmidrule(lr){2-14}
        & nDCG@1 & \cellcolor{gray!10}19.98 & 26.13 & \textbf{28.40} & \cellcolor{gray!10}18.90 & 27.21 & \textbf{29.82} & \cellcolor{gray!10}11.66 & 25.39 & \textbf{28.00} & \cellcolor{gray!10}\textbf{44.83}  & 36.86 & 39.62  \\
        % & nDCG@3 & \cellcolor{gray!10}21.56 & 32.90 & 36.83 & \cellcolor{gray!10}21.56 & 32.90 & 36.83 & \cellcolor{gray!10}0.91 & 0.88 & 0.84 & \cellcolor{gray!10}0.78 & 0.71 & 0.82  \\
        & nDCG@5 & \cellcolor{gray!10}18.64 & 25.72 & \textbf{28.28} & \cellcolor{gray!10}19.36 & 26.51 & \textbf{28.58} & \cellcolor{gray!10}11.57 & 25.97 & \textbf{27.71} & \cellcolor{gray!10}\textbf{39.76}  & 36.46 & 38.25  \\
        % & nDCG@10 & \cellcolor{gray!10}23.10 & 35.76 & 38.27 & \cellcolor{gray!10}23.10 & 35.76 & 38.27 & \cellcolor{gray!10}0.91 & 0.88 & 0.84 & \cellcolor{gray!10}0.78 & 0.71 & 0.82 \\
        & nDCG@20 & \cellcolor{gray!10}22.22 & 29.73 & \textbf{31.70} & \cellcolor{gray!10}22.12 & 30.27 & \textbf{31.68} & \cellcolor{gray!10}14.84 & 31.40 & \textbf{32.33} & \cellcolor{gray!10}38.95  & 39.89 & \textbf{40.78}  \\
        % & nDCG@50 & \cellcolor{gray!10}25.75 & 39.70 & 41.15 & \cellcolor{gray!10}32.48 & 44.81 & 46.69 & \cellcolor{gray!10}0.91 & 0.88 & 0.84 & \cellcolor{gray!10}0.78 & 0.71 & 0.82 \\
        % & nDCG@100 & \cellcolor{gray!10}25.75 & 39.70 & 41.15 & \cellcolor{gray!10}40.12 & 49.27 & 50.94 & \cellcolor{gray!10}0.91 & 0.88 & 0.84 & \cellcolor{gray!10}0.78 & 0.71 & 0.82  \\
        \cmidrule(lr){2-14}
        & MAP@100 & \cellcolor{gray!10}18.79 & 24.62 & \textbf{26.08} & \cellcolor{gray!10}19.15 & 25.69 & \textbf{26.79} & \cellcolor{gray!10}12.03 & 25.18 & \textbf{25.91} & \cellcolor{gray!10}33.32  & 34.87 & \textbf{35.34}  \\
        \toprule[0.5pt]
        \multirow{7}{*}{TriviaQA}& Top-1 & \cellcolor{gray!10}34.16 & 51.22 & \textbf{53.50} & \cellcolor{gray!10}46.30 & 53.81 & \textbf{55.87} & \cellcolor{gray!10}30.76 & 50.85 & \textbf{52.14} & \cellcolor{gray!10}57.47  & 59.52 & \textbf{60.02}  \\
        % & Top-3 & \cellcolor{gray!10}36.59 & 52.11 & 54.74 & \cellcolor{gray!10}36.59 & 52.11 & 54.74 & \cellcolor{gray!10}0.91 & 0.88 & 0.84 & \cellcolor{gray!10}0.78 & 0.71 & 0.82 \\
        & Top-5 & \cellcolor{gray!10}59.49 & 71.74 & \textbf{71.93} & \cellcolor{gray!10}66.28 & 74.05 & \textbf{74.30} & \cellcolor{gray!10}52.65 & 69.00 & \textbf{69.39} & \cellcolor{gray!10}72.4  & 76.93 & \textbf{77.18}  \\
        % & Top-10 & \cellcolor{gray!10}54.46 & 67.26 & 68.50 & \cellcolor{gray!10}54.46 & 67.26 & 68.50 & \cellcolor{gray!10}0.91 & 0.88 & 0.84 & \cellcolor{gray!10}0.78 & 0.71 & 0.82 \\
        & Top-20 & \cellcolor{gray!10}73.91 & \textbf{79.02} & 78.98 & \cellcolor{gray!10}76.41 & \textbf{80.24} & 80.08 & \cellcolor{gray!10}67.18 & 75.48 & \textbf{75.51} & \cellcolor{gray!10}79.77  & 82.44 & \textbf{82.53} \\
        % & Top-50 & \cellcolor{gray!10}0.86 & 0.86 & 0.86 & \cellcolor{gray!10}72.74 & 76.87 & 76.87 & \cellcolor{gray!10}0.91 & 0.88 & 0.84 & \cellcolor{gray!10}0.78 & 0.71 & 0.82 &  \\
        \cmidrule(lr){2-14}
        & nDCG@1 & \cellcolor{gray!10}34.16 & 51.22 & \textbf{53.50} & \cellcolor{gray!10}46.30 & 53.81 & \textbf{55.87} & \cellcolor{gray!10}30.76 & 50.85 & \textbf{52.14} & \cellcolor{gray!10}57.47  & 59.52 & \textbf{60.02}  \\
        % & nDCG@3 & \cellcolor{gray!10}21.56 & 32.90 & 36.83 & \cellcolor{gray!10}21.56 & 32.90 & 36.83 & \cellcolor{gray!10}0.91 & 0.88 & 0.84 & \cellcolor{gray!10}0.78 & 0.71 & 0.82  \\
        & nDCG@5 & \cellcolor{gray!10}30.46 & 47.26 & \textbf{48.61} & \cellcolor{gray!10}41.60 & 51.36 & \textbf{52.45} & \cellcolor{gray!10}27.78 & 48.40 & \textbf{49.22} & \cellcolor{gray!10}49.69  & \textbf{59.21} & 56.57  \\
        % & nDCG@10 & \cellcolor{gray!10}23.10 & 35.76 & 38.27 & \cellcolor{gray!10}23.10 & 35.76 & 38.27 & \cellcolor{gray!10}0.91 & 0.88 & 0.84 & \cellcolor{gray!10}0.78 & 0.71 & 0.82 \\
        & nDCG@20 & \cellcolor{gray!10}31.78 & 48.02 & \textbf{48.82} & \cellcolor{gray!10}40.68 & 52.37 & \textbf{52.71} & \cellcolor{gray!10}29.25 & 50.54 & \textbf{51.05} & \cellcolor{gray!10}46.33  & 56.52 & \textbf{56.67}  \\
        % & nDCG@50 & \cellcolor{gray!10}25.75 & 39.70 & 41.15 & \cellcolor{gray!10}32.48 & 44.81 & 46.69 & \cellcolor{gray!10}0.91 & 0.88 & 0.84 & \cellcolor{gray!10}0.78 & 0.71 & 0.82 \\
        % & nDCG@100 & \cellcolor{gray!10}25.75 & 39.70 & 41.15 & \cellcolor{gray!10}40.12 & 49.27 & 50.94 & \cellcolor{gray!10}0.91 & 0.88 & 0.84 & \cellcolor{gray!10}0.78 & 0.71 & 0.82  \\
        \cmidrule(lr){2-14}
        & MAP@100 & \cellcolor{gray!10}26.61 & 41.77 & \textbf{42.77} & \cellcolor{gray!10}34.85 & 46.82 & \textbf{46.86} & \cellcolor{gray!10}24.02 & 44.33 & \textbf{44.80} & \cellcolor{gray!10}39.40  & 50.74 & \textbf{50.88}  \\
        \bottomrule[1pt]
    \end{tabular}
     \end{threeparttable}
    }
    \vspace{-3mm}
    \caption{Re-ranking results on the test set of datasets of the Top-100 retrieved documents with the GPT-Neo-2.7B model. The best results are highlighted in bold.} \label{tab:neo-rerank}
    \vspace{-3mm}
\end{table*}

\begin{table} [t]
% \addtolength{\tabcolsep}{-0.45pt}
\small
\centering
\resizebox{\columnwidth}{!}{%
\begin{tabular}{lcccccc}
 \toprule
 % \textbf{Model} & \multicolumn{6}{c}{\textbf{LLaMA2-B}} & \multicolumn{6}{c}{\textbf{Mistral-7B}} & \multicolumn{6}{c}{\textbf{Gemma-7B}}\\
    % \cmidrule(lr){2-7} \cmidrule(lr){8-13} \cmidrule(lr){14-19} 
   & \multicolumn{2}{c}{\textbf{NQ}} & \multicolumn{2}{c}{\textbf{WebQ}} & \multicolumn{2}{c}{\textbf{TriviaQA}}  \\
    &  EM & F1 & EM & F1 & EM & F1 \\
% \midrule
\midrule
\multicolumn{7}{c}{\textit{LLaMA2-13B}} \\
\midrule
\rowcolor{gray!10}Contriever & 22.02 & 29.11  &  19.69 & 30.21 & 49.90  &  57.08 \\
\phantom{0000}+ Inference with UPR         & 28.56 & 36.81  &  21.80 & 33.18 & 59.06 & 67.23 \\
\phantom{0000}+ Inference with $\mathrm{UR^3}$         & \textbf{29.00} & \textbf{37.39}  &  \textbf{22.54} & \textbf{34.47} & \textbf{59.43} & \textbf{67.69} \\
\midrule
\rowcolor{gray!10}BM25 & 20.20 & 27.08  &  16.39 & 26.60 & 55.21  &  62.91\\
\phantom{0000}+ Inference with UPR         & 27.62 & 35.63  &  19.59 & 30.25 & 62.25 & 70.27 \\
\phantom{0000}+ Inference with $\mathrm{UR^3}$         & \textbf{29.06} & \textbf{36.82}  &  \textbf{20.67} & \textbf{31.81} & \textbf{62.50} & \textbf{70.59} \\
\midrule
\rowcolor{gray!10}MSS & 19.86 & 26.16  &  16.83  & 27.94 & 49.29  &  56.03\\
\phantom{0000}+ Inference with UPR         & 26.70 & 34.61  &  20.77 & 31.58 & 57.94 & 65.94 \\
\phantom{0000}+ Inference with $\mathrm{UR^3}$         &  \textbf{27.95} & \textbf{35.75}  &  \textbf{21.80} & \textbf{32.99} & \textbf{58.23} & \textbf{66.20} \\
\midrule
\rowcolor{gray!10}DPR & 30.30 & \textcolor{red}{\textbf{38.42}}  &  \textcolor{red}{\textbf{22.79}}  & 34.36 & 55.33  &  62.96 \\
\phantom{0000}+ Inference with UPR         & 30.86 & 37.93  &  21.78 & 33.74 & 60.61 & 68.79  \\
\phantom{0000}+ Inference with $\mathrm{UR^3}$         & \textbf{30.97} & 38.38  &  22.75 & \textbf{35.50} & \textbf{60.98} & \textbf{69.21} \\
\midrule
\multicolumn{7}{c}{\textit{Mistral-7B}} \\
\midrule
\rowcolor{gray!10}Contriever & 20.69 & 26.61 & 14.37 & 24.39 &49.89 & 56.94 \\
\phantom{0000}+ Inference with UPR         & 25.29 & 32.30 & 16.78 & 26.67 & 59.35 & 67.01 \\
\phantom{0000}+ Inference with $\mathrm{UR^3}$         & \textbf{25.93} & \textbf{32.98} & \textbf{17.42} & \textbf{27.77} & \textbf{59.32} & \textbf{67.13}  \\
\midrule
\rowcolor{gray!10}BM25 & 19.14 & 25.31 & 13.29 & 23.03 & 54.91 & 62.15\\
\phantom{0000}+ Inference with UPR         & 24.32 & 31.10 & 15.55 & 25.30 & 62.50 & 69.88\\
\phantom{0000}+ Inference with $\mathrm{UR^3}$         & \textbf{25.37} & \textbf{32.24} & \textbf{15.85} & \textbf{25.71} & \textbf{62.61} & \textbf{70.02} \\
\midrule
\rowcolor{gray!10}MSS & 18.20 & 24.28 & 13.98 & 23.79 & 48.41 & 55.22\\
\phantom{0000}+ Inference with UPR         &  24.04 & 30.88 & 16.04 & 26.24 & 58.09 & 65.76\\
\phantom{0000}+ Inference with $\mathrm{UR^3}$         & \textbf{24.27} & \textbf{31.03} & \textbf{16.54} & \textbf{26.64} & \textbf{58.11} & \textbf{65.86} \\
\midrule
\rowcolor{gray!10}DPR &  28.17 & 34.9 & 18.21 & 28.55 & 55.03 & 62.24 \\
\phantom{0000}+ Inference with UPR         &  26.65 & 34.01 & 17.42 & 27.90 & 60.53 & 68.54 \\
\phantom{0000}+ Inference with $\mathrm{UR^3}$         & \textbf{28.20} & \textbf{34.94} & \textbf{18.40} & \textbf{28.57} & \textbf{60.69} & \textbf{68.74}  \\
\midrule
\multicolumn{7}{c}{\textit{Gemma-7B}} \\
\midrule
\rowcolor{gray!10}Contriever &17.40 & 25.13 & 14.71 & 26.05 & 45.54 & 53.66 \\
\phantom{0000}+ Inference with UPR         &  22.11 & 30.70 & 15.00 & 26.95 & 55.78 & 64.67\\
\phantom{0000}+ Inference with $\mathrm{UR^3}$         &  \textbf{23.02} & \textbf{31.27} & \textbf{15.60} & \textbf{27.04} & \textbf{55.89} & \textbf{64.96}\\
\midrule
\rowcolor{gray!10}BM25 & 16.40 & 23.84 & 12.35 & 22.84 & 52.34 & 60.89\\
\phantom{0000}+ Inference with UPR         & 22.52 & 31.09 & 14.12 & \textbf{25.22} & 59.52 & 68.23\\
\phantom{0000}+ Inference with $\mathrm{UR^3}$         & \textbf{22.99} & \textbf{31.45} & \textbf{14.42} & 25.05 & \textbf{59.66} & \textbf{68.32}\\
\midrule
\rowcolor{gray!10}MSS & 14.38 & 21.44 & 12.20 & 22.91 & 43.56 & 51.54\\
\phantom{0000}+ Inference with UPR         & 20.94 & 29.40 & 14.35 & \textbf{26.79} & \textbf{54.42} & 63.08\\
\phantom{0000}+ Inference with $\mathrm{UR^3}$         &  \textbf{22.11} & \textbf{30.22} & \textbf{14.76} & 26.58 & 54.30 & \textbf{63.28}\\
\midrule
\rowcolor{gray!10}DPR & \textcolor{red}{\textbf{24.43}} & \textcolor{red}{\textbf{33.14}} & \textcolor{red}{\textbf{16.68}} & 28.00 & 50.76 & 59.59\\
\phantom{0000}+ Inference with UPR         &  23.43 & 32.44 & 17.22 & 28.54 & 57.27 & 66.16\\
\phantom{0000}+ Inference with $\mathrm{UR^3}$         & 24.10 & 33.07 & 16.44 & \textbf{28.02} & \textbf{57.30} & \textbf{66.21}\\
 \bottomrule
\end{tabular}
}
\caption{EM and F1 scores for the open-domain QA task. We perform inference with the re-ranked Top-1 results of Table~\ref{tab:mistral-rerank}. The best performing models are highlighted in bold. We highlight the best scores obtained by original retriever in red.}
\label{tab:odqa-mistral}
% \label{tab:odqa-llama2-7b}
%\vspace{-6pt}
\end{table}

\begin{table} [t]
% \addtolength{\tabcolsep}{-0.45pt}
\small
\centering
\resizebox{\columnwidth}{!}{%
\begin{tabular}{lcccccc}
 \toprule
 % \textbf{Model} & \multicolumn{6}{c}{\textbf{LLaMA2-B}} & \multicolumn{6}{c}{\textbf{Mistral-7B}} & \multicolumn{6}{c}{\textbf{Gemma-7B}}\\
    % \cmidrule(lr){2-7} \cmidrule(lr){8-13} \cmidrule(lr){14-19} 
   & \multicolumn{2}{c}{\textbf{NQ}} & \multicolumn{2}{c}{\textbf{WebQ}} & \multicolumn{2}{c}{\textbf{TriviaQA}}  \\
    &  EM & F1 & EM & F1 & EM & F1 \\
% \midrule
\midrule
\multicolumn{7}{c}{\textit{Top-1}} \\
\midrule
\rowcolor{gray!10}Contriever & 15.90 & 22.00  &  14.42 & 24.46 & 40.31  & 47.67 \\
\phantom{0000}+ Inference with UPR         & 20.97 & 27.90  & 15.26 & 25.10 & 52.16 & 60.40 \\
\phantom{0000}+ Inference with $\mathrm{UR^3}$         & \textbf{21.93} & \textbf{29.06}  &  \textbf{15.45} & \textbf{24.97} & \textbf{51.90} & 60.34 \\
\midrule
\rowcolor{gray!10}BM25 & 15.65 & 21.38  &  12.55 & 21.55 & 48.41  &  56.60\\
\phantom{0000}+ Inference with UPR         & 20.75 & 27.71  &  14.47 & 24.59 & 55.68 & 64.22\\
\phantom{0000}+ Inference with $\mathrm{UR^3}$         & \textbf{21.75} & \textbf{28.91}  &  \textbf{15.70} & \textbf{25.27} & \textbf{56.73} & \textbf{65.13} \\
\midrule
\rowcolor{gray!10}MSS & 13.60 & 19.34  &  11.81  & 21.63 & 39.98  &  47.17\\
\phantom{0000}+ Inference with UPR         & 19.78 & 26.98  &  14.76 & 24.52 & 50.45 & 58.75\\
\phantom{0000}+ Inference with $\mathrm{UR^3}$         & \textbf{21.69} & \textbf{28.66}  &  \textbf{15.06} & \textbf{25.27} & \textbf{51.20} & \textbf{59.31} \\
\midrule
\rowcolor{gray!10}DPR & 23.38 & 30.69  &  16.20  & 26.28 & 46.86  & 54.87 \\
\phantom{0000}+ Inference with UPR         & 22.13 & 29.58  & 15.26 & 25.18 & 53.84 & 62.17 \\
\phantom{0000}+ Inference with $\mathrm{UR^3}$         & \textbf{24.29} & \textbf{31.16}  &  \textbf{17.03} & \textbf{27.13} & \textbf{54.08} & \textbf{62.40} \\
\midrule
\multicolumn{7}{c}{\textit{Top-3}} \\
\midrule
\rowcolor{gray!10}Contriever & 14.93 & 20.36  &  12.16 & 22.03 & 40.23  &  49.53 \\
\phantom{0000}+ Inference with UPR         & 19.31 & 25.51  & 13.44 & 22.65 & 49.74  &  59.38 \\
\phantom{0000}+ Inference with $\mathrm{UR^3}$         & \textbf{19.98} & \textbf{25.77}  &  \textbf{14.03} & \textbf{23.23} & \textbf{50.08} & \textbf{59.66} \\
\midrule
\rowcolor{gray!10}BM25 & 14.35 & 20.04  &  10.73 & 19.76 & 49.46  &  58.58\\
\phantom{0000}+ Inference with UPR         & 19.56 & 25.90  & 12.89 & 21.58 & 56.17 & 65.81\\
\phantom{0000}+ Inference with $\mathrm{UR^3}$         & \textbf{20.02} & \textbf{26.90}  &  \textbf{13.78} & \textbf{22.56} & \textbf{56.55} & \textbf{65.82} \\
\midrule
\rowcolor{gray!10}MSS & 13.63 & 19.48  &  12.45  & 21.53 & 40.36  &  49.44\\
\phantom{0000}+ Inference with UPR         & 18.70 & 24.96  & 13.14 & 22.82 & 48.93 & 58.33\\
\phantom{0000}+ Inference with $\mathrm{UR^3}$         & \textbf{19.47} & \textbf{26.20}  &  \textbf{13.93} & \textbf{23.01} & \textbf{49.21} & \textbf{58.51} \\
\midrule
\rowcolor{gray!10}DPR & 19.61 & 26.21  &  13.48  & 22.55 & 44.79  &  54.11\\
\phantom{0000}+ Inference with UPR         & 20.42 & 27.19  &  13.24 & 21.88 & \textbf{51.55} &\textbf{61.24} \\
\phantom{0000}+ Inference with $\mathrm{UR^3}$         & \textbf{22.08} & \textbf{29.24}  &  \textbf{15.35} & \textbf{24.16} & 51.54 & 61.23 \\
\midrule
\multicolumn{7}{c}{\textit{Top-5}} \\
\midrule
\rowcolor{gray!10}Contriever & 18.50 & 23.80  &  13.29 & 23.08 & 46.53  &  55.12 \\
\phantom{0000}+ Inference with UPR         & 22.33 & 28.81  &  15.21 & 24.14 & \textbf{55.43} & 64.10 \\
\phantom{0000}+ Inference with $\mathrm{UR^3}$         & \textbf{22.55} & \textbf{28.82}  &  \textbf{15.21} & \textbf{24.28} & 55.36 & \textbf{64.26} \\
\midrule
\rowcolor{gray!10}BM25 & 16.45 & 22.05  & 12.01 & 20.69 & 54.96  & 63.46\\
\phantom{0000}+ Inference with UPR         & 21.91 & 28.28  &  13.88 & 22.84 & 60.83 & 69.51\\
\phantom{0000}+ Inference with $\mathrm{UR^3}$         & \textbf{22.27} & \textbf{28.70}  &  \textbf{14.57} & \textbf{23.27} & \textbf{60.90} & \textbf{69.51} \\
\midrule
\rowcolor{gray!10}MSS & 16.59 & 22.22  &  13.39  & 22.99 & 46.35  &  54.79\\
\phantom{0000}+ Inference with UPR         & 20.72 & 26.87  & 14.22 & 23.30 & 54.04 & 62.55\\
\phantom{0000}+ Inference with $\mathrm{UR^3}$         & \textbf{21.27} & \textbf{27.68}  &  \textbf{14.27} & \textbf{23.57} & \textbf{54.81} & \textbf{62.62} \\
\midrule
\rowcolor{gray!10}DPR & 22.60 & 28.84  &  13.63  & 22.21 & 50.61  &  58.98 \\
\phantom{0000}+ Inference with UPR         & 23.74 & 30.21  &  15.65 & 24.25 & 56.53 & 65.22 \\
\phantom{0000}+ Inference with $\mathrm{UR^3}$         & \textbf{24.54} & \textbf{30.96}  &  \textbf{16.34} & \textbf{24.72} & \textbf{56.63} & \textbf{65.29} \\
 \bottomrule
\end{tabular}
}
\caption{EM and F1 scores for the open-domain QA task with different number of input documents on the LLaMA2-7B model. The best performing models are highlighted in bold.}
\label{tab:odqa-llama2-7b}
%\vspace{-6pt}
\end{table}
% This is an appendix.
\end{document}